\documentclass[journal]{IEEEtran}
\usepackage{amsmath,amsfonts}
\usepackage{algorithmic}
\usepackage{algorithm}
\usepackage{array}
\usepackage[caption=false,font=normalsize,labelfont=sf,textfont=sf]{subfig}
\usepackage{textcomp}
\usepackage{stfloats}
\usepackage{url}
\usepackage{verbatim}
\usepackage{graphicx}
\usepackage{cite}
\usepackage[citecolor=blue, colorlinks]{hyperref}
\usepackage{colortbl}  
\usepackage{xcolor}
\usepackage{caption}
\usepackage{graphicx}
\usepackage{float} 
\usepackage{color}
\usepackage{soul}
\usepackage{booktabs}
\usepackage{multirow}
\usepackage[normalem]{ulem}
\useunder{\uline}{\ul}{}
\usepackage{bbding}

\newcommand{\revise}[1]{\textcolor{black}{#1}}

\begin{document}

\title{ ResFlow: Fine-tuning Residual Optical Flow for Event-based High Temporal Resolution Motion Estimation}

\author
{
Qianang Zhou,
Zhiyu Zhu,
Junhui Hou,~\IEEEmembership{Senior Member,~IEEE},
Yongjian Deng,
Youfu Li,~\IEEEmembership{Fellow,~IEEE},
Junlin Xiong,~\IEEEmembership{Member,~IEEE}


\thanks{Qianang Zhou is with the Department of Automation, University of Science and Technology of China, Anhui 230026, China, and is also with the Department of Computer Science, City University of Hong Kong, Hong Kong (email: qianazhou2-c@my.cityu.edu.hk).}

\thanks{Zhiyu Zhu and Junhui Hou are with the Department of Computer Science, City University of Hong Kong, Hong Kong (email:zhiyuzhu2-c@my.cityu.edu.hk; jh.hou@cityu.edu.hk)}

\thanks{Youfu Li is with the Department of Mechanical Engineering, City University of Hong Kong, Hong Kong (email:meyfli@cityu.edu.hk)}

\thanks{Yongjian Deng is with the College of Computer Science, Beijing University of Technology, Beijing, China (yjdeng@bjut.edu.cn)}

\thanks{Junlin Xiong is with the Department of Automation, University of Science and Technology of China, Anhui 230026, China (email:xiong77@ustc.edu.cn)}
}

\markboth{Revised Manuscript submitted to IEEE}%
{Shell \MakeLowercase{\textit{et al.}}: A Sample Article Using IEEEtran.cls for IEEE Journals}


\maketitle

\begin{abstract}
\revise{Event cameras offer great potential for high-temporal-resolution (HTR) motion estimation in dynamic real-world scenarios. However, the lack of dense HTR ground truth in real-world datasets prevents fully leveraging the high temporal resolution potential of event cameras. Furthermore, the intrinsic sparsity of event data introduces additional challenges for optimization and supervision.}
\revise{To address these issues, we propose a residual-based paradigm that decomposes HTR optical flow into a global linear component and high-frequency residuals.}
The residual paradigm effectively mitigates the impacts of event sparsity on optimization and is compatible with any LTR algorithm. 
\revise{In addition, to bridge the supervision gap caused by the lack of HTR ground truth, we incorporate novel learning strategies.}
Specifically, we initially employ a shared refiner to estimate the residual flows, enabling both LTR supervision and HTR inference. Subsequently, we introduce regional noise to simulate the residual patterns of intermediate flows, facilitating the adaptation from LTR supervision to HTR inference. Additionally, we show that the noise-based strategy supports in-domain self-supervised training. 
Comprehensive experimental results demonstrate that our approach achieves state-of-the-art accuracy among existing HTR methods, highlighting its effectiveness and superiority.
\end{abstract}

\begin{IEEEkeywords}
event-based vision, optical flow, deep learning.
\end{IEEEkeywords}

\section{Introduction}

\IEEEPARstart{O}{ptical} flow serves as a cornerstone in numerous computer vision tasks, capturing the motion of pixels over time. Recent advancements in learning-based approaches have significantly improved optical flow estimation~\cite{dosovitskiy2015flownet,sun2018pwcnet}. Frameworks~\cite{teed2020raft,jiang2021gma} incorporating all-pairs correlation computation and iterative refinement have emerged as pivotal breakthroughs, achieving remarkable performance across diverse scenarios. 
Despite these advancements, significant challenges remain, particularly in handling high dynamic range scenes and coping with the frame-rate limitations of conventional cameras, which hinder motion perception under complex conditions.

Event cameras present a promising solution to these limitations~\cite{gallego2020event}. Unlike traditional cameras, event cameras operate asynchronously, with individual pixels independently detecting changes in brightness. This design enables ultra-low latency and an exceptionally high dynamic range, making event cameras particularly effective in extreme lighting conditions and rapid motion. These advantages position event cameras as a powerful tool for motion estimation across a broader range of real-world scenarios. By leveraging the complementary strengths of event and frame-based cameras, substantial progress has been achieved in tasks such as motion deblurring~\cite{liu2023motiondeblur, zhu2023learningdeblur, zhu2022learning, chen2025event}, image reconstruction~\cite{nie2020highhtr, yu2025event}, and object tracking~\cite{wang2023visevent, zhu2024crsot}. Additionally, event-based approaches have demonstrated competitive performance compared to traditional frame-based methods~\cite{chen2022ecsnetstf, chen2024segment,xie2024eventstf,zhu2024continuous,deng2021mvfll}. However, the potential of event cameras for high-frequency applications remains underexplored. The inherently high temporal resolution of event cameras offers an unprecedented opportunity for estimating continuous motion with high precision, paving the way for transformative advancements in various downstream applications~\cite{liu2024video, liu2022edflow, zhu2023cross, wu2024motion, chen2024crossei, zhu2025spatio, shi2024polarity, liu2024event}. \revise{Consequently, high-temporal-resolution optical flow emerges as a promising direction in event-based vision, potentially redefining motion analysis in dynamic and challenging environments. Related applications have also been explored in several works~\cite{khan2022non, cao2025decomposition, khan2021quantum, moshayedi2023designing, moshayedi2024design, moshayedi2024evaluating, r2r1, r2r2, r2r3, r2r4}.}

\revise{A key bottleneck in this field is the lack of HTR ground truth in real-world event-based datasets.}
The domain gap between synthetic datasets and real-world scenarios further limits the practical applicability of existing algorithms. Several methods have been proposed to address challenges in HTR optical flow estimation, broadly classified into two categories: (\textit{i}) self-supervised methods, which use motion compensation losses to guide the estimation of intermediate motion, and (\textit{ii}) cumulative methods, which progressively accumulate HTR flows to form LTR flows, indirectly supervising HTR flows with LTR ground truth. Self-supervised methods~\cite{paredes2023taming} heavily depend on contrast maximization loss~\cite{gallego2018unifying,zhu2018evflownet}, leading to performance that significantly lags behind supervised approaches. In contrast, cumulative methods~\cite{ye2023towards, ponghiran2023sequential} suffer from the absence of explicit constraints on intermediate flows, resulting in substantial error accumulation and ultimately limiting their effectiveness. 
\revise{The sparsity and noise inherent to event data are further amplified at higher temporal resolutions, making HTR flow estimation particularly challenging.}

Estimating intermediate flows from scratch typically requires globally robust all-pairs similarity volumes, yet the extreme sparsity and noise of HTR event feature undermine this assumption. On the other hand, LTR algorithms address event sparsity by aggregating event information over the entire duration, achieving high accuracy for LTR optical flow. This raises a critical question: \textit{Can we provide robust references for HTR flows and transform HTR flow prediction into residual prediction, to overcome the trade-off between frequency and accuracy?}

To address this challenge, we propose a two-stage residual-based paradigm for HTR optical flow estimation and a corresponding training strategy based on LTR ground truth. Unlike cumulative methods, our approach predicts the residual flow between linear and nonlinear motion.
By requiring only local robustness in correlation features, residual prediction helps mitigate the adverse effects of event sparsity and noise at high temporal resolutions.
Additionally, our training strategy resolves the discrepancies between LTR and HTR residuals, enabling effective supervision using LTR ground truth. The proposed residual-based framework addresses the inherent trade-off between temporal resolution and accuracy in cumulative methods, offering a novel and robust solution for event-based HTR optical flow estimation.

Specifically, we propose a novel residual-based framework that supports LTR supervision and HTR optical flow estimation.
Linear motion derived from the global stage provides a robust and accurate reference for refining nonlinear motion. Residual flow prediction across varying temporal spans is unified and implemented via a shared residual refiner, enabling both LTR supervision and HTR optical flow estimation. To facilitate effective supervision of HTR flow predictions using LTR ground truth, we introduce two strategies: optical flow velocity transformation and noise-based training. In particular, we introduced regional noise that emulates residual flow patterns, facilitating the adaptation from LTR to HTR residual flows. With these learning strategies, our method utilizes ground truth with a frequency of 10 Hz for supervision, while performing inference at a frequency $15\times$ higher (150 Hz). 
Extensive experiments validate that our algorithm achieves state-of-the-art performance across all existing HTR methods, in terms of both end-point error and flow-warp loss.
Comprehensive ablation studies further validate the effectiveness of the proposed strategies. 

In summary, our key contributions are:

\begin{itemize}
    \item A residual-based framework for HTR optical flow estimation that decomposes HTR flow into an LTR component and an HTR residual, effectively mitigating the adverse effects of event sparsity.
    \item A novel LTR ground truth-based training strategy that integrates velocity transformation and a noise-based training pipeline, enabling LTR supervision while achieving HTR inference.
    \item Comprehensive evaluation: An improved warping method enhances the reliability of motion compensation, and thorough ablation studies validate the effectiveness of the proposed approaches.
\end{itemize}

The rest of this paper is organized as follows: Sec.~\ref{sec:rewo} reviews frame-based optical flow and event-based LTR and HTR optical flow. Sec.~\ref{sec:preliminary} introduces the preliminaries and event representations used in this study. Sec.~\ref{sec:method} details the proposed framework and train strategy for transitioning from LTR supervision to HTR inference. Sec.~\ref{sec:exp} analyzes the experimental results. Finally, Sec.~\ref{sec:conclusion} concludes the paper.

\section{Related Work}
\label{sec:rewo}

\subsection{Optical Flow Estimation}
Data-driven methods have achieved remarkable progress in optical flow estimation~\cite{dosovitskiy2015flownet, sun2018pwcnet, teed2020raft, hui2018liteflownet, jiang2021gma, xu2022gmflow}. FlowNet~\cite{dosovitskiy2015flownet} pioneered end-to-end optical flow prediction, demonstrating the potential of learning-based approaches for this task. Building on this, PWC-Net~\cite{sun2018pwcnet} introduced an efficient architecture that combines pyramid processing, warping, and cost-volume construction. By leveraging multi-scale information, PWC-Net improved the ability to capture diverse motion magnitudes, thereby enhancing both accuracy and efficiency. However, these methods rely on local correlation computations, which struggle with large motions. RAFT~\cite{teed2020raft} addressed this limitation by constructing a 4D all-pairs correlation volume and refining flow predictions iteratively through recurrent neural networks, establishing a robust framework for subsequent methods~\cite{jiang2021gma, huang2022flowformer}. Building on RAFT's 4D correlation volume, GMFlow~\cite{xu2022gmflow} redefined optical flow estimation as a global matching problem, offering an efficient alternative for optical flow prediction. Most recently, diffusion-model-based approaches~\cite{saxena2024surprising} have achieved promising success in the field of optical flow. Our framework aligns with the correlation-based paradigm, with the residual flow strategy drawing inspiration from RAFT's iterative refinement module.

\subsection{Event-based Optical Flow}
Current approaches to event-based flow estimation can be broadly categorized into two types: LTR estimation, aligned with the frame rate of traditional cameras, and HTR estimation, which significantly surpasses the frame rates achievable by conventional cameras.

\vspace{0.5em}
\textbf{Low Temporal Resolution.}
Event-based LTR optical flow methods primarily adapt architectures successfully in frame-based optical flow. For instance, EV-FlowNet~\cite{zhu2018evflownet} voxelizes events into frame-like representations and employs the FlowNet~\cite{dosovitskiy2015flownet} architecture for self-supervised flow estimation. As RAFT~\cite{teed2020raft} gained prominence for frame-based optical flow, E-RAFT~\cite{gehrig2021eraft} adapted its design to event-based vision. Inspired by PWC-Net~\cite{sun2018pwcnet}, IDNet~\cite{wu2024idnet} iteratively warps events and estimates optical flow, avoiding the computational overhead of constructing correlation volumes. Some methods exploit the unique characteristics of events, such as their high temporal resolution and spatial sparsity. ECDDP~\cite{yang2025ecddp} proposed a self-supervised framework for dense prediction tasks, leveraging large-scale training on synthetic datasets. TMA~\cite{liu2023tma} integrates intermediate motion information based on the correlation architecture to achieve high-quality LTR optical flow estimation. In addition to data-driven methods, model-based approaches have been explored. MultiCM~\cite{shiba2022multicm, shiba2024secrets}, for instance, warps events along optical flow trajectories and formulates an energy function based on the image of warped events to handle complex scenarios. LTR methods integrate intermediate motion cues to improve the estimation accuracy of the overall trajectory, which our approach leverages as a robust reference for HTR prediction.

\begin{figure*}[t]
  \centering
  \includegraphics[width=0.99\linewidth]{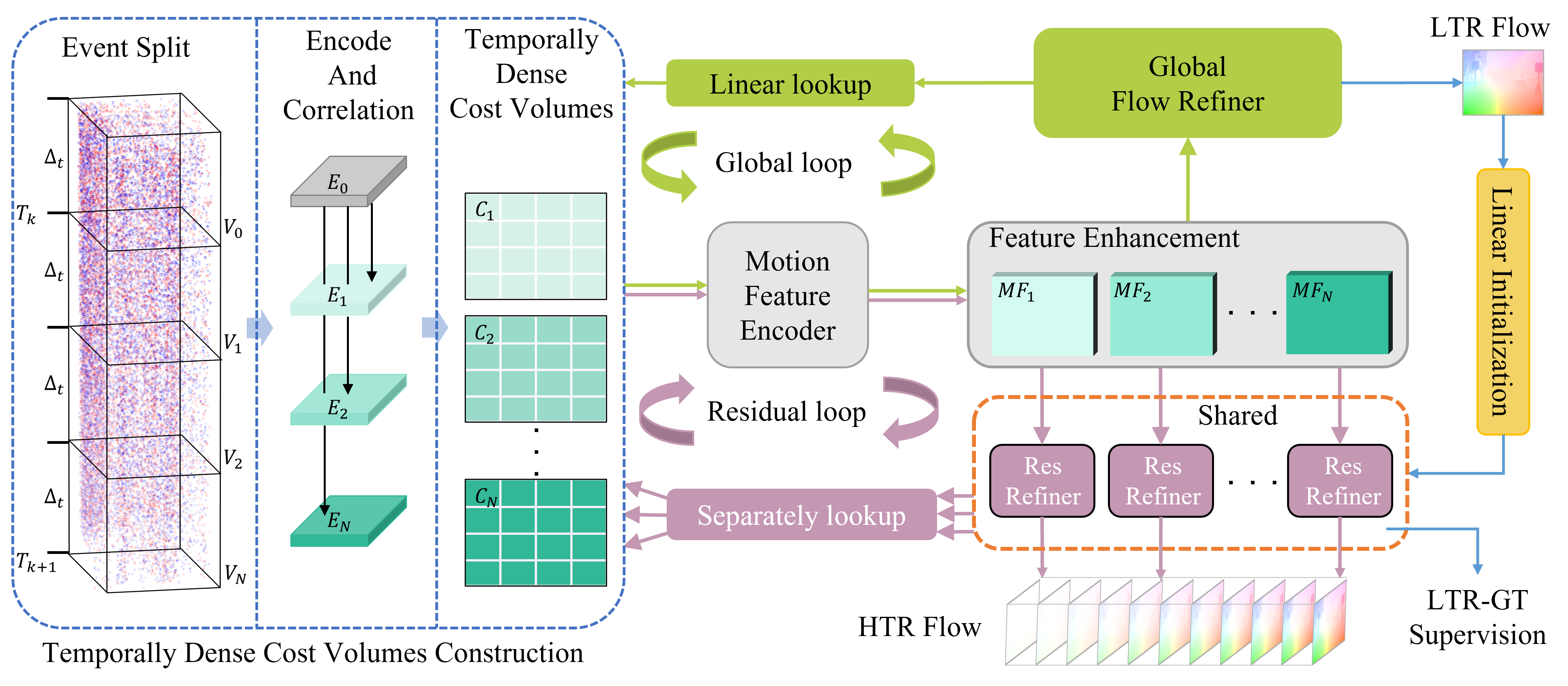}
  \caption{Our HTR residual flow estimation framework. HTR optical flow estimation is decoupled into two stages. We begin by splitting events, extracting features, and computing correlations to construct the temporally dense cost volumes, which serve as a shared foundation for both stages. In the global stage, all intermediate motion features are aggregated to estimate accurate LTR optical flow. 
  \revise{In the residual stage, the HTR residual flow at each intermediate time is predicted using its corresponding motion features.}
  The shared use of cost volumes, motion encoders, and feature enhancement modules across both stages greatly reduces the model complexity. Additionally, by sharing the residual refiner across different intermediate times, the algorithm supports LTR supervision and HTR inference.}
  \label{fig:framework}
\end{figure*}

\vspace{0.5em}
\textbf{High Temporal Resolution.} 
High-frequency event data enables HTR optical flow estimation, but the lack of HTR ground truth in real-world datasets poses significant challenges. Some algorithms impose self-supervised constraints on intermediate motion. For example, Hagenaars et al.~\cite{hagenaars2021nips} replaced artificial neural networks with spiking neural networks to asynchronously estimate optical flow, supervised by the average timestamp loss~\cite{zhu2018evflownet}. TCM~\cite{paredes2023taming} computes the average timestamp loss at multiple temporal scales to improve the robustness of self-supervised learning. Continuous Flow~\cite{hamann2024motion} employs contrast maximization loss~\cite{gallego2018unifying} to supervise sparse point trajectories. Algorithms supervised with LTR ground truth often adopt an accumulation-based paradigm for implicit supervision of intermediate flows. EVA-Flow~\cite{ye2023towards} estimates HTR flow incrementally by propagating warped intermediate features for subsequent predictions. Ponghiran et al.~\cite{ponghiran2023sequential} reformulate event-based optical flow estimation as a sequential learning problem, embedding intermediate flow information into the model's hidden states. Synthetic datasets provide HTR ground truth for some works. DCEIFlow~\cite{wan2022dceiflow} generates pseudo-second-frame features by combining event features with the first frame and applying the iterative refinement framework. BFlow~\cite{gehrig2024bflow} models long-term pixel trajectories as Bezier curves, predicting control points using a RAFT-based architecture. Unlike cumulative approaches, the proposed residual-based framework avoids challenging optimization and implicit supervision. Coupled with the novel learning strategy, ResFlow supports LTR supervision and HTR inference.

\section{Preliminary and Event Representation}\label{sec:preliminary}
Event cameras record changes in pixel intensity asynchronously. An event $e_i$ is triggered when the logarithmic intensity change at a pixel exceeds the threshold $C$. Each event $e_i$ includes a timestamp $t$, spatial coordinates $(x_i,y_i)$, and a polarity $p$, where $p=+1$ indicates an increase in intensity, and $p=-1$ represents a decrease. Due to the extremely high temporal resolution of events, they are often discretized into several temporal bins for processing~\cite{zhu2018evflownet}. Specifically, events are embedded into a 3D grid representation $\mathbf{V}$ with $B$ bins as follows:
\begin{equation}
\begin{split}
t_i^* &= (B-1)(t_i-t_1)/(t_N-t_1),\\
\mathbf{V}(x,y,t) &= \sum\limits_{i}p_ik_b(x-x_i)k_b(y-y_i)k_b(t-t_i^*),\\
k_b(a)&=\max(0,1-|a|),
\end{split}
\end{equation}
where $k_b(\cdot)$ is a bilinear interpolation function.

In this study, to construct correlation volumes for intermediate flow, we split the event stream in the interval of $[T_k, T_{k+1}]$ into a series of target segments $\{\mathbf{V}_1, \mathbf{V}_2, ..., \mathbf{V}_N\}$ uniformly, each with an average time interval of $\Delta_t$. To estimate the optical flow from $T_k$ to $T_{k+1}$, we use the segment from time interval of $[T_k-\Delta_t , T_k]$ as the reference $\mathbf{V}_0$, following previous methods~\cite{liu2023tma,gehrig2024bflow}. Eventually, we split the event into a reference $\mathbf{V}_0$ and multiple intermediate targets $\{\mathbf{V}_n, n\in[1,N]\}$ to estimate the high-temporal-resolution optical flow from $T_k$ to $T_{k+1}$.

\section{Proposed Method}
\label{sec:method}

Estimating intermediate flows is challenging due to the inherent sparsity and noise of event features. To address this issue and reduce the training burden, we decompose HTR optical flow into global motion component and residual flows, as described in Sec.~\ref{sec:framework}. However, the lack of HTR Ground Truth (GT) remains an issue. In Sec.~\ref{sec:training}, we analyze the differences between HTR and LTR residuals and design a novel training process to facilitate the adaptation of the network from LTR supervision to HTR inference.

\subsection{HTR Flow Estimation via Learning HTR Residual}\label{sec:framework}
To mitigate the impact of event sparsity and noise on HTR prediction, our framework is structured in two stages, as shown in Fig.~\ref{fig:framework}. In the global stage, the LTR optical flow is predicted using the overall event information, providing a robust reference for the intermediate flows. 
\revise{The residual stage focuses on estimating nonlinear residual flows at intermediate times using temporally local motion features.}
We proceed to detail the two stages individually.

\vspace{0.5em}
\noindent\textbf{Global Stage}. By decomposing the HTR optical flow into LTR components and HTR residuals, our framework supports various LTR algorithms. However, to maintain computational efficiency, we adopt temporally dense correlation-based LTR methods~\cite{liu2023tma, gehrig2024bflow}, which enable the reuse of intermediate computations during residual refinement. As done in Sec.~\ref{sec:preliminary}, we split the event into a reference $\mathbf{V}_0$ and multiple intermediate targets $\{\mathbf{V}_n,n\in[1,N]\}$ to estimate the LTR optical flow from $T_k$ to $T_{k+1}$, where $\mathbf{V}_0$ (resp. $\mathbf{V}_N$) corresponds to the segment with timestamp of $T_k$ (resp. $T_{k+1}$). \textcolor{black}{All event segments $\{V_n,n\in[1,N]\}$ are processed by the share-weights encoder to extract features $\{E_n,n\in[1,N]\}$. Afterward, the all-pairs pixel similarity is generated between each target and the reference:}

\begin{equation}
        \mathbf{C}_n=\frac{\mathbf{E}_0\mathbf{E}_n^T}{\sqrt{D}}, n\in \{1,2,...,N\},
\end{equation}
where the $\mathbf{C}_n \in \mathbb{R}^{H W\times H W}$ is the cost volume between the reference $\mathbf{E}_0$ and the $n$-th target $\mathbf{E}_n$, and the $D$ is the channel of feature $\mathbf{E}$.
The temporally dense cost volumes $\{\mathbf{C}_n,n\in[1,N]\}$ are shared across both stages.

\textcolor{black}{Throughout the iterations of the global stage, we perform a linear lookup within temporally dense cost volumes to retrieve the similarity cost, which is further used to derive motion features for refining the flow estimation. Given the estimated optical flow $\mathbf{F}^{(j)}$ for the $j$-th iteration, the corresponding temporal linear lookup is formulated as follows:}

\begin{equation}
\begin{split}
    \mathbf{F}_n^{(j)} = \frac{n\Delta_t}{T_{k+1}-T_k}\cdot \mathbf{F}^{(j)}, n\in\{1,2,...,N\},\\
\end{split}
\end{equation}
where the $\mathbf{F}_{n}^{(j)}$ is the linear optical flow at the $n$-th target after $j$-th iteration, $\Delta_t$ is the discretization time interval between neighbor voxels. The intermediate flow cost is retrieved from $\mathbf{C}_n$ according to $\mathbf{F}_n$ and further encoded as a motion feature. Subsequently, following the previous method~\cite{liu2023tma}, we employ attention-based enhancement of the motion features and aggregate the enhanced features into a unified motion representation for LTR flow update.

\begin{figure}[t]
  \centering
  \includegraphics[width=0.79\linewidth]{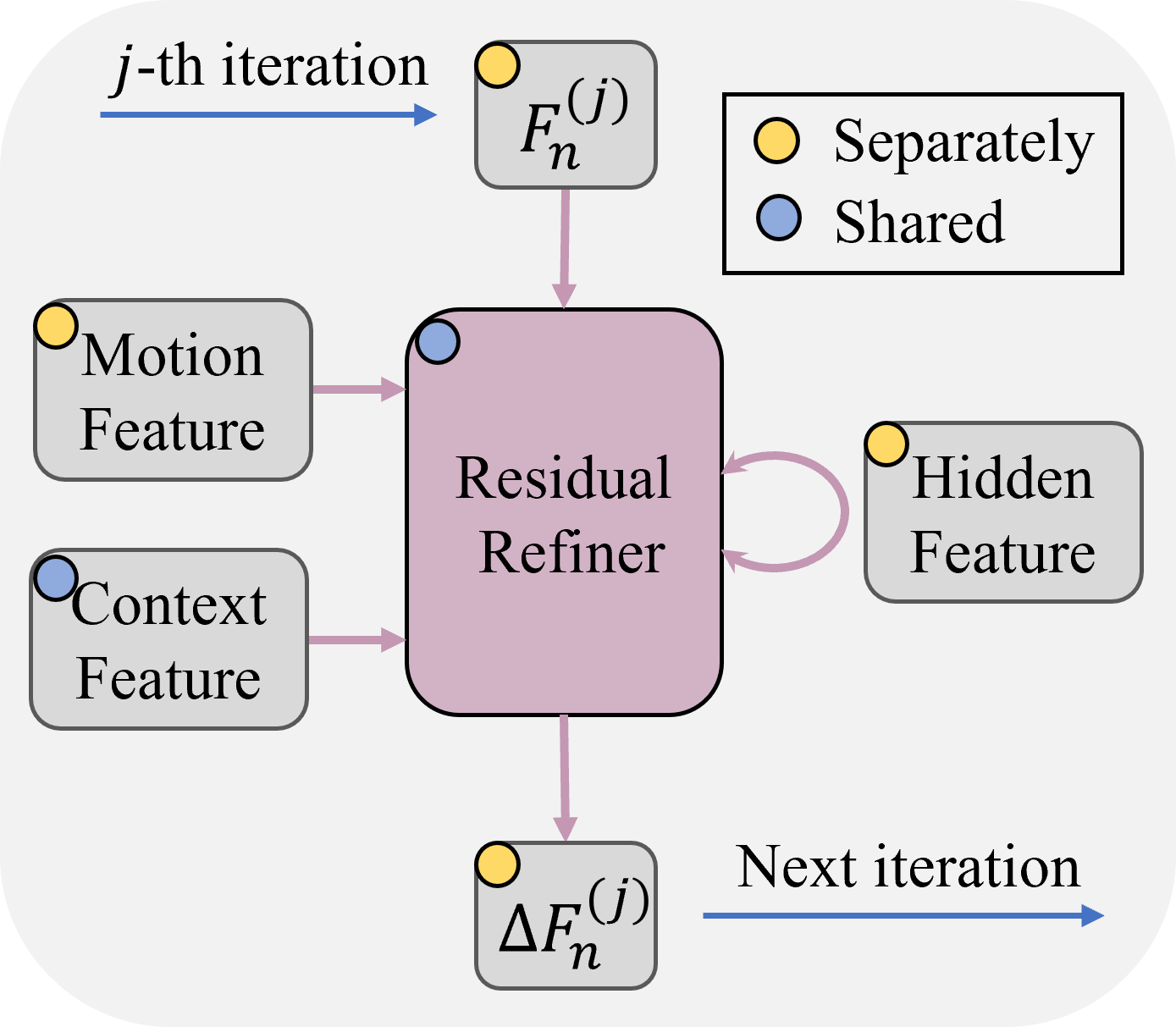}
  \caption{Shared structure of the residual refiner in Fig.~\ref{fig:framework}. The residual refiner employs shared parameters to support both LTR supervision and HTR inference. Context features are shared across all intermediate flows, while optical flow and motion features are specific to each intermediate time step. The refiner is based on ConvGRU, whose hidden features are uniquely associated with each time step.}
  \label{fig:rfr}
\end{figure}

\begin{figure*}[t]
  \centering
  \subfloat[Conceptual demonstration of residual flow.]{\includegraphics[width=0.39\linewidth]{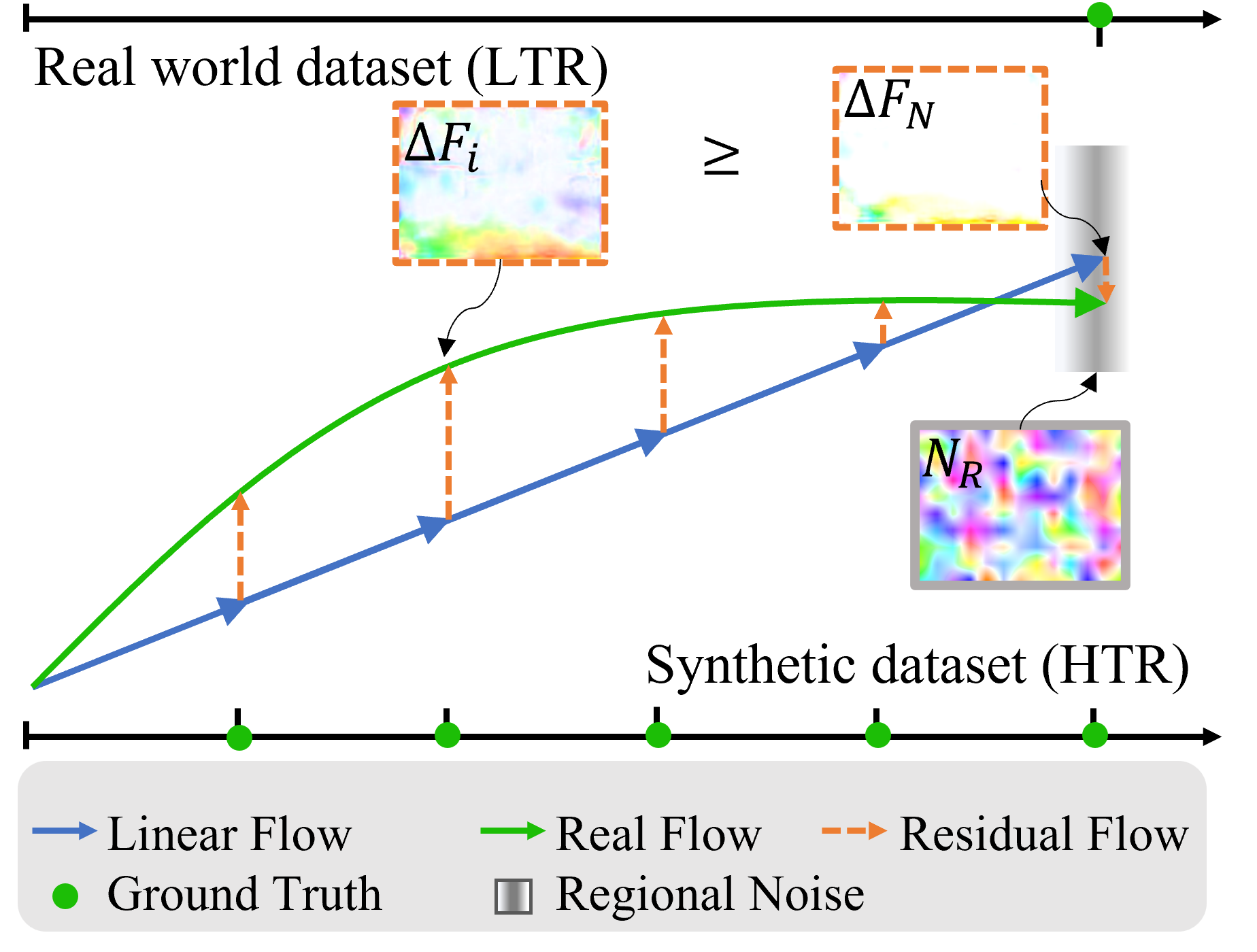}}
  \subfloat[Residual flow and noises patterns.]{\includegraphics[width=0.59\linewidth]{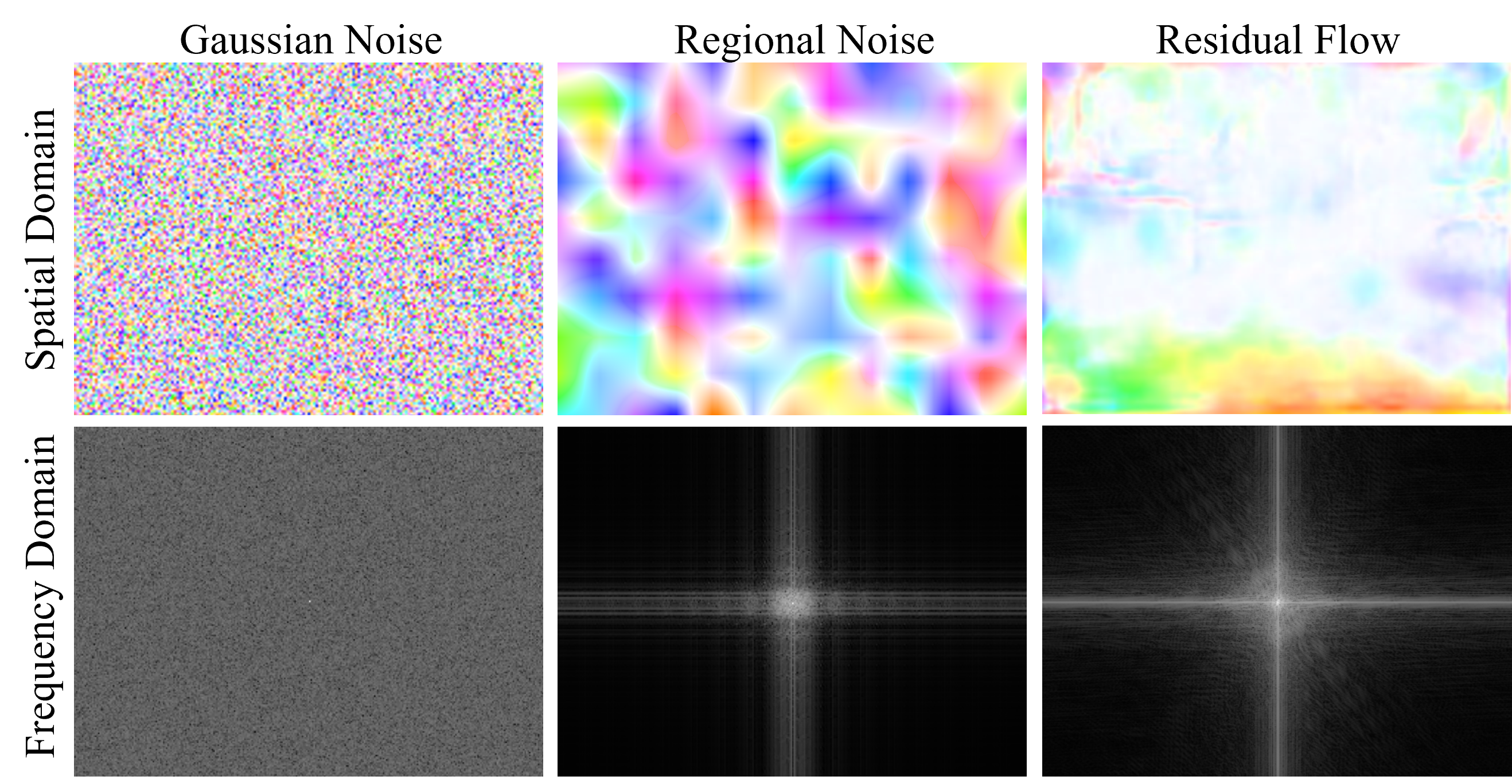}}
  \caption{Illustration of the motivation behind the proposed noise injection training strategy. (a) The linearly interpolated optical flow (represented by the blue line) exhibits a significant discrepancy compared to the \textcolor{green}{GT flow} (depicted by the green line). As supervision is limited to the LTR flow, addressing the substantial gap between the estimated and GT flows can be effectively achieved by introducing perturbations to align the errors with those at intermediate points. 
   (b) demonstrates the proposed perturbation strategy, showcasing visualizations of pure Gaussian noise, regional noise, and residual flow. Both spatial and frequency domain visualizations reveal that the regional noise perturbation strategy effectively replicates the distribution of residual flow. In contrast, traditional Gaussian noise fails to capture the residual patterns accurately.}
  \label{fig:resfconcept}
\end{figure*}

\vspace{0.5em}
\noindent \textbf{Residual Stage.} 
We estimate the HTR flow based on the temporally interpolated LTR flow in a residual manner. Fig.~\ref{fig:resfconcept} illustrates the concept of residual flow, which represents the difference between nonlinear motion and the initialized linear motion. To estimate the nonlinear motion trajectory (green flow), we initialize the intermediate moments (blue flows) using the global motion predicted by the LTR model and subsequently estimate the corresponding residual flow (orange flows).

\textcolor{black}{To achieve this, we design a shared-weights residual refiner tailored for residual estimation at different timestamps, as shown in Fig.~\ref{fig:rfr}. The refiner utilizes shared context features and parameters while keeping the motion and flow features specific to each timestamp. This design ensures that all intermediate flows are refined by a unified model and guided by consistent context. Additionally, the refiner provides two key benefits: (a) it allows training with LTR ground truth, and (b) it enables residual estimation at any intermediate timestamp, given the motion features and initialized flow. These properties allow for effective supervision of HTR flow using limited LTR ground truth and provide explicit constraints compared to previous implicit methods~\cite{ye2023towards,ponghiran2023sequential}.}

\textcolor{black}{ResFlow achieves high efficiency in both training and inference. As illustrated in Fig.~\ref{fig:framework}, the residual flow is estimated at discrete timestamps within temporally dense cost volumes, avoiding the computational overhead of recomputing base features for each timestamp. The motion encoder and feature enhancement module are shared across both stages, eliminating the need for retraining. During the residual stage, only the residual refiner requires training. The accurate LTR flow initialized in the global stage reduces the number of iterations needed for residual refinement, enabling efficient HTR flow estimation with minimal iterations.}

\subsection{Scale-consistent HTR Learning Strategy} \label{sec:training}
Synthetic datasets provide HTR-GT, enabling strong supervision for all intermediate flows. In contrast, real-world datasets only offer LTR-GT, as illustrated in Fig.~\ref{fig:resfconcept}. Benefiting from the shared parameter design, our framework supports training with LTR-GT. However, a critical challenge remains: how to bridge the gap between HTR and LTR residual predictions to ensure that LTR supervision can effectively generalize to HTR predictions.

We revisited the residual flow prediction process and identified two key differences between intermediate and final residual flow predictions. The \textbf{first} difference lies in the time-scale dependency of optical flow, where flows corresponding to different temporal spans vary significantly in magnitude. Specifically, the magnitudes of intermediate flows are consistently smaller than those of LTR flows. Since optical flow magnitude often serves as a critical feature for refinement, this discrepancy hinders the residual refiner's adaptability to intermediate flows. The \textbf{second} difference arises from the magnitude of residual flows. As illustrated in Fig.~\ref{fig:resfconcept}, the global stage output closely approximates the LTR GT, resulting in minimal residuals. During training, the small residual provides weak supervision signals, which are insufficient for residual learning. Moreover, residuals associated with nonlinear and linear motion are predominantly distributed across intermediate flows. As depicted in the figure, the intermediate residual flow $\Delta \mathbf{F}_{i}$ is notably larger than the final residual flow $\Delta\mathbf{F}_N$, which further diminishes the effectiveness of LTR supervision. To address these discrepancies, we propose adaptation learning strategies that effectively generalize LTR residuals to HTR residuals.

\begin{figure*}[t]
\centering
  \includegraphics[width=0.99\linewidth]{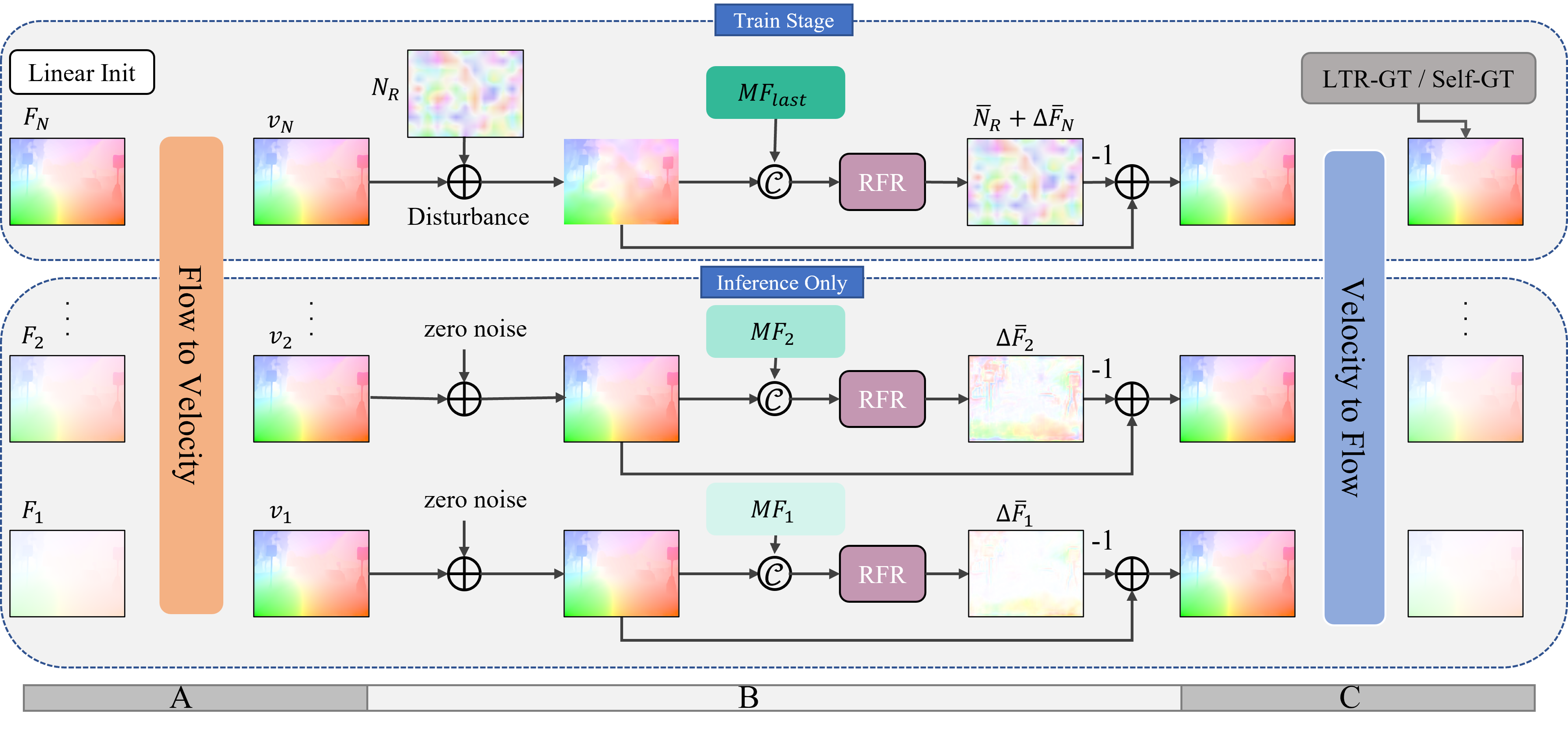}
  \caption{\textbf{Training and inference pipeline illustration.} We propose a strategy to support LTR supervision and HTR inference, promoting the generalization of LTR supervision to HTR predictions, which can be divided into the following three parts. \textbf{Region A}: Optical flow is transformed into a scale-invariant average velocity, addressing magnitude disparities between flows across varying temporal spans. \textbf{Region B}: The proposed regional noise is incorporated during LTR training to model HTR residual patterns. Here, -1 indicates that the estimated noise is subtracted from the noise-perturbed flow. Finally, in \textbf{Region C}: The average velocity is converted to optical flow for lookup and output. The conversion between optical flow and velocity is still necessary during inference, but noise is no longer required. Moreover, under the noise strategy, the small LTR residual $\Delta\mathbf{F}_N$ enables self-supervised residual training, where the LTR-GT is replaced by the initialized LTR optical flow $\mathbf{F}_N$.}
  \label{fig:trainfw}
\end{figure*}

\vspace{0.5em}
\noindent\textbf{Velocity Transformation.} To address the first difference, we propose replacing displacement estimation with scale-invariant velocity estimation. Taking $T_{k+1}-T_k$ as the unit time, for an intermediate time $T_k+n\Delta_t$, the network estimates the average velocity from time $T_k$ to $T_k+n\Delta_t$. Unlike optical flow, velocity is scale-invariant, resulting in minimal differences between LTR and HTR. Therefore, we reformulate the optical flow estimation problem as an average velocity estimation. Given the average velocity for the entire duration as initialization, the network estimates the average velocity from time $T_k$ to $T_k+n\Delta_t$ based on the motion features at $T_k+n\Delta_t$. As shown in Fig.~\ref{fig:trainfw}, we introduce a flow and velocity converter in the residual stage. The conversion between optical flow and velocity is defined as:
\begin{equation}
        \mathbf{v}_{n}=\mathbf{F}_{n} \cdot\frac{T_{k+1}-T_k}{n\cdot \Delta_t}, n\in\{1,2,...,N\},
\end{equation}
where the $\mathbf{v}_n$ is the average velocity between the reference and the $n$-th targets, $\mathbf{F}_n$ is the corresponding flow, the $T_{k+1}-T_k$ is the time interval of the LTR flow, and also the unit time for velocity.

We replace the optical flow estimate with the average velocity to fill the gap caused by different time intervals. As illustrated in Fig.~\ref{fig:trainfw}, the average velocities $\{v_n,n\in[1,~N]\}$ of different intermediate flows have similar magnitudes, which simplifies the inference for intermediate flows. By predicting scale-invariant average velocities instead of optical flow displacements, we effectively address the first discrepancy between LTR and HTR residual estimation. Notably, the average velocity is ultimately transformed back into optical flow for lookup and output.

\vspace{0.5em}
\noindent\textbf{Regional Noise Training.} The input flow features exhibit a consistent temporal scale in the average velocity setting. To further mitigate the effects of amplitude differences in the residual flow, we propose adding artificial perturbations under LTR supervision. As illustrated in Fig.~\ref{fig:trainfw}, random noise $\mathbf{N}_R$ is added to the initial flow $\mathbf{F}_N$. This perturbation not only enlarges the LTR residuals but also introduces randomness, preventing the network from overfitting LTR residuals. The motivation for adding noise is further clarified in Fig.~\ref{fig:resfconcept}. The magnitude of the residual $\Delta\mathbf{F}_n$ is too small to enable effective supervision, making it challenging for the residual refiner to learn the necessary corrections to the optical flow. The introduced noise will be predicted as part of the residual, compels the residual refiner to adjust the optical flow under significant perturbations.

With the motivation for adding noise established, we propose that the introduced noise should exhibit a pattern similar to that of the residual flow. The residual flow arises from the difference between linear and nonlinear motion in the scene and is typically regional in nature. Simple Gaussian noise cannot effectively model this regional residual. Therefore, we propose a spatially correlated noise that more closely aligns with the residual flow pattern. Specifically, we generate a low-resolution Gaussian noise and then upsample it to the original resolution. The \textbf{regional noise} $\mathbf{N}_R$ is synthesized as follows:
\begin{equation}
    \mathbf{N}_R = Up(\mathbf{G},S), \mathbf{G}\sim \mathcal{N}(\mathbf{0},\mathbf{I}),
\end{equation}
where the $Up(\cdot)$ is the bilinear upsampling function, $\mathbf{G}\in\mathbb{R}^{\frac{H}{S}\times\frac{W}{S}\times 2}$ is the low resolution gaussian noise, and the $S$ is the scaling factor.

Compared to Gaussian noise, regional noise exhibits spatial correlation. As illustrated in Fig.~\ref{fig:resfconcept}, we compare the two types of noise with the residual flow in both spatial and frequency domains. The results indicate that Gaussian noise contains higher frequencies, whereas regional noise has more low-frequency components, making it closer to the residual flow pattern. In Sec.~\ref{sec:exp}, we train with Gaussian noise and regional noise as interference, respectively, to validate our analysis.

In summary, the training and inference pipeline of the residual refiner is depicted in Fig~\ref{fig:trainfw}. During training, regional noise is added to the initial flow, and the corrected outputs of the residual refiner are supervised using the LTR-GT. Notably, the training of the residual refiner can also be conducted in a self-supervised manner. Since the LTR residual is generally small enough to be ignored, residual refiner primarily focuses on correcting the artificial perturbations. As shown in Fig~\ref{fig:trainfw}, the initialization flow can be used for self-supervision, even when LTR-GT is unavailable. The effectiveness of this approach will be demonstrated in Sec.~\ref{sec:ablation}.

Training is performed only at LTR flow where the GT is available, while inference can be performed at any intermediate time step. Given computational constraints, we perform inference only at the target time points. Our framework supports inference up to $15 \times$ the frequency of LTR algorithms, with results presented and analyzed in Sec.~\ref{sec:ablation}.

\subsection{Training Objectives}
Supervision is performed only where the LTR ground truth is available, which is the last flow $\mathbf{F}_N$. 
\revise{We follow the standard setup of correlation-based methods and adopt the $L_1$ distance between the predictions and the ground truth as the loss function, due to its robustness to outliers and stable gradient properties. The supervision is applied to each output of the iterator.}
\begin{equation}
\mathcal{L}_1=\sum\limits_{j=1}^{m}\gamma^{m-j}\|\mathbf{\hat{F}}_N^{(j)}-\mathbf{F}_{gt}\|_1,
\end{equation}
where the $m$ is the total number of residual refiner iterations, the $\mathbf{\hat{F}}_N^{(j)}$ is the output of the $j$-th iteration, and $\gamma$ is the decay factor. For self-supervision, we replace $\mathbf{F}_{gt}$ with the LTR optical flow predicted by the global stage.

To further enhance the constraints on residual prediction, we perform the lookup using the nonlinear flow obtained after the residual stage and supervise the corresponding LTR optical flow, $\mathbf{\hat{F}}$:
\begin{equation}
\mathcal{L}_2=\|\mathbf{\hat{F}}-\mathbf{F}_{gt}\|_1.
\end{equation}
The final loss consists of two components:
\begin{equation}
\mathcal{L}=\mathcal{L}_1+\mathcal{L}_2.
\end{equation}
During the training of the residual stage, the parameters of the global stage are frozen to stabilize the LTR initialization. As a result, the output of the global stage is not supervised.

\section{Experiment}
\label{sec:exp}

\subsection{Dataset and Metrics.}
\vspace{0.5em}
\noindent\textbf{Dataset and Setup.}
To ensure fair comparisons with prior methods, we conducted extensive experiments on the real-world dataset DSEC-Flow~\cite{gehrig2021dsec}. This dataset encompasses a diverse range of driving scenarios, including challenging conditions such as nighttime, sunrise, sunset, and tunnels. It provides an official training set and a publicly accessible online benchmark.

In our experiments, the global-stage parameters were initialized using pre-trained LTR models~\cite{liu2023tma} and remained frozen during the residual-stage training. Training utilized all DSEC data with available LTR-GT, and evaluation was performed on the official test set. The number of intermediate targets, $N$, was set to 5, with an LTR prediction frequency of 10 Hz and an HTR frequency of 50 Hz. During testing, we increased the ResFlow frequency to 150 Hz without retraining, and results are detailed in Sec.~\ref{sec:ablation}. The residual refiner iteration count was set to 4, the regional noise downsampling factor $S$ was set to 6, and the probability of adding noise was set to 0.6.

\vspace{0.5em}
\noindent\textbf{LTR Metrics.} The LTR metrics were derived by comparing the final flow with the LTR-GT, including the End-Point-Error (EPE) and the outlier ratio (\%Out). EPE was calculated over all valid GT points, while outliers were defined as points with an error exceeding 3 pixels from the GT. These metrics were computed online using DSEC-Flow's official benchmark.

\vspace{0.5em}
\noindent\textbf{HTR Metrics.} We evaluate the accuracy of HTR optical flow trajectories using the FWL metric, which is widely employed to assess continuous flow and HTR flow~\cite{hamann2024motion, shiba2022multicm, ye2023towards, paredes2023taming}. Given a set of events $\mathcal{E}$ and the estimated motion trajectory $\mathbf{F}$, the Image of Warped Events (IWE) is generated by warping the events back to the initial time along $\mathbf{F}$. The FWL value is defined as the variance of IWE relative to that of the identity warp:
\begin{equation}
    FWL:=\frac{\sigma^2(\mathbf{I}(\mathcal{E},\mathbf{F}))}{\sigma^2(\mathbf{I}(\mathcal{E},0))},
\end{equation}
where $\sigma^2(\cdot)$ represents the variance function and $\mathbf{I}$ denotes the IWE. FWL reflects the accuracy of motion compensation, with higher trajectory accuracy producing greater contrast in the IWE.

Motion compensation involves converting forward optical flow to backward optical flow, a task for which no exact solution exists. Previous works~\cite{shiba2022multicm,ye2023towards} proposed coarse approximations that often result in suboptimal compensation. In ERAFT~\cite{gehrig2021eraft}, a flow propagation method was introduced to propagate flow in its motion direction, which was used to initialize the optical flow of the next frame. We adopt this method for forward-to-backward flow conversion. Specifically, this method propagates initial values along the optical flow direction and averages them at the endpoints. Given the forward optical flow $\mathbf{F}_{i\rightarrow i+1}$, the backward flow $\mathbf{F}_{i+1\rightarrow i}$ is computed as:
\begin{equation}
    \begin{split}
        g(x_i)&=x_i+\mathbf{F}_{i\rightarrow i+1}(x_i),\\
        \mathbf{F}_{i+1\rightarrow i}&=-\frac{\sum_{\forall x_i}k_b(x-g(x_i))\mathbf{F}_{i\rightarrow i+1}(x_i)}{\sum_{\forall x_i}k_b(x-g(x_i))},
    \end{split}
\end{equation}
where $g(x_i)$ represents the position of the current pixel in the next frame, and $k_b(\cdot)$ denotes the bilinear interpolation function. 

\begin{figure}
  \includegraphics[width=0.99\linewidth]{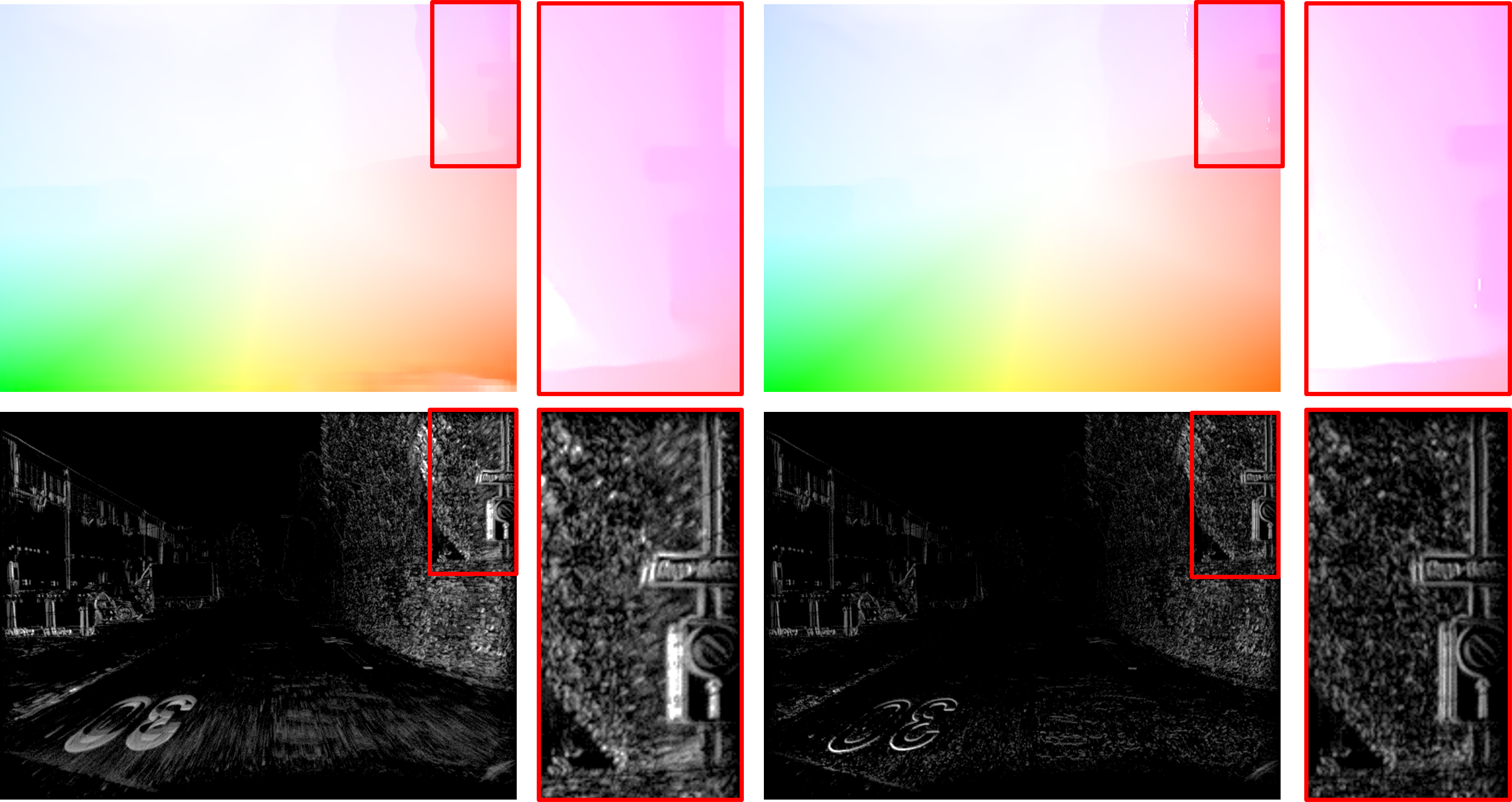}
  \caption{\textbf{Comparison of warping methods}. The top row presents backward optical flow generated by different algorithms, while the bottom row displays the corresponding Images of Warped Events (IWEs). The left column shows results from previous warping methods~\cite{shiba2022multicm}, and the right column depicts outcomes from the improved warping method. The enhanced backward optical flow aligns the scene structure more accurately with motion, ensuring better consistency with the event stream. This improvement significantly enhances motion compensation, resulting in sharper edges in the IWE.}
  \label{fig:warpcomp}
\end{figure}

In our experiments, the propagation-based conversion method significantly outperformed prior straightforward approaches~\cite{shiba2022multicm}. Fig.~\ref{fig:warpcomp} illustrates a qualitative comparison between them. Backward optical flow comparisons show that the scene structure produced by the propagation-based approach adapts to motion, aligning with the event stream positions. This alignment enhances warping accuracy and substantially improves the IWE quality.

\begin{table*}[t]
\centering
\caption{\textbf{DSEC-Flow evaluation results}. The \colorbox[gray]{0.9}{gray} column highlights the HTR metric. \textbf{Bold} indicates the best result in HTR methods, \underline{underline} indicates the second best. `Sup.' indicates whether the method is supervised with LTR-GT. `*' denotes the straightforward warp method~\cite{shiba2022multicm}. ``$\uparrow$" (``$\downarrow$") indicates the larger (resp. smaller), the better.}
\begin{tabular}{@{}clccc >{\columncolor[gray]{0.9}}c cc >{\columncolor[gray]{0.9}}c cc >{\columncolor[gray]{0.9}}c cc>{\columncolor[gray]{0.9}}cc@{}}
\toprule
\toprule
\multicolumn{2}{c}{\multirow{2}{*}{Method}} & \multirow{2}{*}{Sup.} & \multicolumn{3}{c}{Overall}                   & \multicolumn{3}{c}{interlaken\_00\_b}         & \multicolumn{3}{c}{interlaken\_01\_a}         & \multicolumn{3}{c}{thun\_01\_a}               \\ \cmidrule(l){4-15} 
\multicolumn{2}{c}{}                        &                       & EPE$\downarrow$           & \%Out$\downarrow$          & FWL$\uparrow$            & EPE$\downarrow$            & \%Out$\downarrow$          & FWL$\uparrow$           & EPE$\downarrow$            & \%Out$\downarrow$          & FWL$\uparrow$           & EPE$\downarrow$            & \%Out$\downarrow$          & FWL$\uparrow$           \\ \midrule
\multirow{5}{*}{LTR}    & ERAFT*~\cite{gehrig2021eraft}            & \Checkmark                     & 0.79          & 2.68          & 1.33          & 1.39          & 6.19          & 1.42          & 0.90          & 3.91          & 1.56          & 0.65          & 1.87          & 1.30          \\
                        & MultiCM~\cite{shiba2022multicm}           & \XSolidBrush                    & 3.47          & 30.86         & 1.37          & 5.74          & 38.93         & 1.46          & 3.74          & 31.37         & 1.63          & 2.12          & 17.68         & 1.32          \\
                        & EVFlowNet~\cite{zhu2018evflownet}         & \XSolidBrush                    & 3.86          & 31.45         & 1.30          & 6.32          & 47.95         & 1.46          & 4.91          & 36.07         & 1.42          & 2.33          & 20.92         & 1.32          \\
                        & TMA*~\cite{liu2023tma}              & \Checkmark                     & 0.75          & 2.39          & 1.60          & 1.35          & 5.60          & 1.65          & 0.84          & 3.35          & 1.78          & 0.61          & 1.61          & 1.54          \\
                        & TMA~\cite{liu2023tma}               & \Checkmark                     & 0.75          & 2.39          & 2.07          & 1.35          & 5.60          & 2.34          & 0.84          & 3.35          & 2.53          & 0.61          & 1.61          & 1.80          \\ \midrule
\multirow{4}{*}{HTR}    & TCM-S~\cite{paredes2023taming}             & \XSolidBrush                    & 9.66          & 86.44         & {\ul 1.91}    & 9.86          & 87.24         & {\ul 1.89}    & 9.33          & 86.70         & {\ul 2.07}    & 8.71          & 86.45         & {\ul 1.81}    \\
                        & TCM-M~\cite{paredes2023taming}             & \XSolidBrush                    & {\ul 2.33}          & 17.77         & 1.26          & 3.34          & 25.72         & 1.33          & 2.49          & 19.15         & 1.40          & 1.73          & 10.39         & 1.21          \\
                        & ContFlow~\cite{hamann2024motion}    & \XSolidBrush                    &  3.20     & {\ul 15.21}   & 1.46          & {\ul 3.21}    & {\ul 20.45}   & 1.58          & {\ul 2.38}    & {\ul 17.40}    & 1.70           & {\ul 1.39}    & {\ul 7.36}    & 1.30           \\
                        & \textbf{Ours}     & \Checkmark                     & \textbf{0.75} & \textbf{2.50} & \textbf{2.14} & \textbf{1.36} & \textbf{5.96} & \textbf{2.43} & \textbf{0.85} & \textbf{3.41} & \textbf{2.65} & \textbf{0.62} & \textbf{1.69} & \textbf{1.82} \\ \midrule \\ \midrule
\multicolumn{2}{c}{\multirow{2}{*}{Method}} & \multirow{2}{*}{Sup.} & \multicolumn{3}{c}{thun\_01\_b}               & \multicolumn{3}{c}{zurich\_city\_12\_a}       & \multicolumn{3}{c}{zurich\_city\_14\_c}       & \multicolumn{3}{c}{zurich\_city\_15\_a}       \\ \cmidrule(l){4-15} 
\multicolumn{2}{c}{}                        &                       & EPE$\downarrow$           & \%Out$\downarrow$          & FWL$\uparrow$            & EPE$\downarrow$            & \%Out$\downarrow$          & FWL$\uparrow$           & EPE$\downarrow$            & \%Out$\downarrow$          & FWL$\uparrow$           & EPE$\downarrow$            & \%Out$\downarrow$          & FWL$\uparrow$           \\ \midrule
\multirow{5}{*}{LTR}    & ERAFT*~\cite{gehrig2021eraft}            & \Checkmark                     & 0.58          & 1.52          & 1.25          & 0.61          & 1.06          & 0.91          & 0.71          & 1.91          & 1.47          & 0.59          & 1.30          & 1.40          \\
                        & MultiCM~\cite{shiba2022multicm}           & \XSolidBrush                    & 2.48          & 23.56         & 1.28          & 3.86          & 43.96         & 1.08          & 2.72          & 30.53         & 1.44          & 2.35          & 20.99         & 1.39          \\
                        & EVFlowNet~\cite{zhu2018evflownet}         & \XSolidBrush                    & 3.04          & 25.41         & 1.33          & 2.62          & 25.80         & 1.03          & 3.36          & 36.34         & 1.24          & 2.97          & 25.53         & 1.33          \\
                        & TMA*~\cite{liu2023tma}              & \Checkmark                     & 0.56          & 1.43          & 1.52          & 0.57          & 0.88          & 1.29          & 0.75          & 3.04          & 1.79          & 0.56          & 1.08          & 1.68          \\
                        & TMA~\cite{liu2023tma}               & \Checkmark                    & 0.56          & 1.43          & 1.91          & 0.57          & 0.88          & 1.31          & 0.75          & 3.04          & 2.10          & 0.56          & 1.08          & 2.20          \\ \midrule
\multirow{4}{*}{HTR}    & TCM-S~\cite{paredes2023taming}             & \XSolidBrush                    & 9.38          & 86.68         & {\underline{1.66}}    & 11.54         & 85.35         & \textbf{1.40} & 10.18         & 86.39         & \textbf{2.50} & 8.54          & 86.30         & {\ul 2.01}    \\
                        & TCM-M~\cite{paredes2023taming}             & \XSolidBrush                    & 1.66          & {\ul 9.34}    & 1.25          & {\ul 2.72}    & 26.65         & 1.04          & 2.64          & 23.01         & 1.38          & 1.69          & 9.98          & 1.23          \\
                        & ContFlow~\cite{hamann2024motion}    & \XSolidBrush                    & {\ul 1.54}    & 9.69          & 1.33          & 8.33          & {\ul 22.39}   & 1.13          & {\ul 1.78}    & {\ul 12.99}   & 1.56          & {\ul 1.45}    & {\ul 8.34}    & 1.51          \\
                        & \textbf{Ours}   & \Checkmark                    & \textbf{0.57} & \textbf{1.49} & \textbf{1.96} & \textbf{0.57} & \textbf{0.91} & {\ul 1.31}    & \textbf{0.76} & \textbf{3.30} & {\ul 2.11}    & \textbf{0.56} & \textbf{1.16} & \textbf{2.26} \\ \bottomrule \bottomrule
\end{tabular}
\label{tab:mainresult}
\end{table*}

\begin{figure*}
  \includegraphics[width=0.99\linewidth]{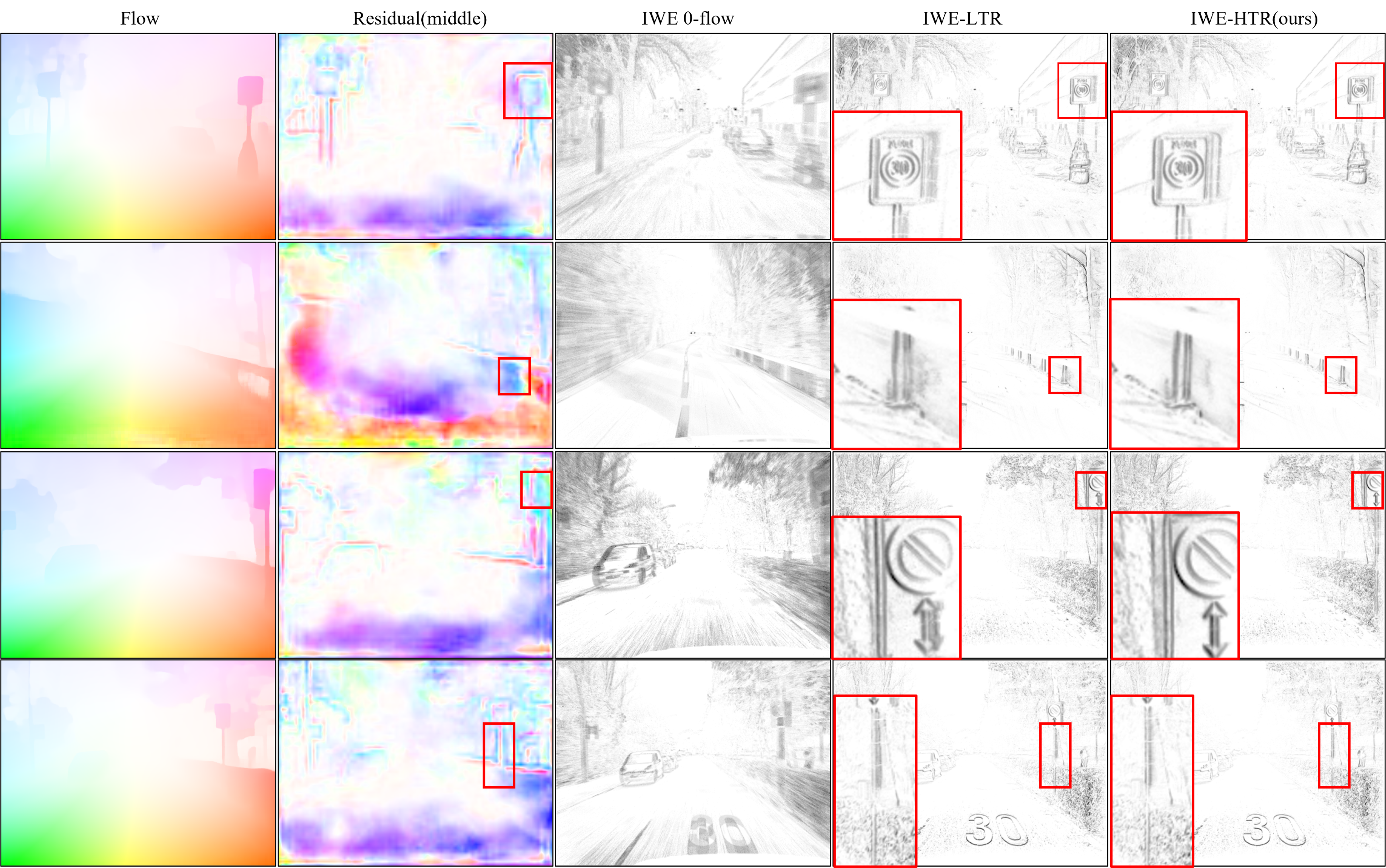}
  \caption{\textbf{Qualitative Results Comparison}. The residual refiner estimates HTR residuals based on LTR linear motion. To evaluate its effectiveness, motion compensation is performed on events using LTR and HTR optical flows, respectively. The results show that our method significantly enhances edge clarity in the IWE. Additionally, residuals at intermediate moments are visualized to analyze the distribution of nonlinear motion regions, with a detailed analysis provided in the main text.}
  \label{fig:mainresult}
\end{figure*}

\subsection{DSEC Dataset}

Table~\ref{tab:mainresult} presents the quantitative evaluation on the DSEC-Flow dataset, while Fig.~\ref{fig:mainresult} shows the qualitative results. To ensure a comprehensive assessment, we report results for both LTR and HTR methods, along with the corresponding LTR (EPE and \%Out) and HTR (FWL) metrics.
Several conclusions can be drawn from these results. First, our proposed warping method, based on flow propagation, improves the reliability of the FWL metric. As two of the most critical evaluation metrics for event-based optical flow, EPE and FWL serve distinct purposes: EPE measures trajectory endpoint errors using LTR-GT, while FWL assesses trajectory accuracy based on IWE contrast. In previous works, these metrics often exhibited significant discrepancies, as shown in Table~\ref{tab:mainresult}. Specifically, unsupervised methods achieved favorable FWL scores but showed substantial endpoint errors, whereas supervised methods performed poorly on FWL. Previous studies~\cite{shiba2022multicm, paredes2023taming} attributed this issue to the lack of precision in warping methods. By incorporating the flow propagation method from ERAFT~\cite{gehrig2021eraft}, we achieved greater consistency between EPE and FWL, leading to a more reliable evaluation. In Table~\ref{tab:mainresult}, we present FWL scores for TMA~\cite{liu2023tma} using both the previous and propagation-based warping methods. The results demonstrate that the propagation-based method significantly enhances FWL scores. This improved method aligns closely with the EPE metric, further highlighting the high quality of DSEC-Flow labeling~\cite{gehrig2021eraft}.


Secondly, we evaluated the performance of ResFlow and compared it to various state-of-the-art methods, as summarized in Table~\ref{tab:mainresult}. These methods are categorized into LTR and HTR approaches, with their respective supervision strategies indicated in the table. Compared to other HTR methods, our approach achieves substantial improvements in both endpoint error and trajectory accuracy. Notably, while most HTR methods utilize IWE contrast for supervision, our method relies exclusively on GT supervision and outperforms existing HTR methods on the FWL metric. For LTR methods, ResFlow maintains low endpoint errors while significantly surpassing others in FWL performance. Overall, ResFlow enhances the trajectory accuracy of linear models while preserving low endpoint errors. It is worth noting that FWL differences below 0.1 are considered significant~\cite{shiba2022multicm}. 
Fig.~\ref{fig:mainresult} provides a qualitative comparison between the LTR baseline and the proposed ResFlow. The LTR model struggles to capture high-frequency nonlinear trajectories, particularly around object edges and fast motions, leading to noticeable blur and distortion artifacts. In contrast, ResFlow enhances motion compensation by accurately predicting HTR residuals, resulting in sharper and more temporally consistent IWE.

\begin{figure}
  \includegraphics[width=0.99\linewidth]{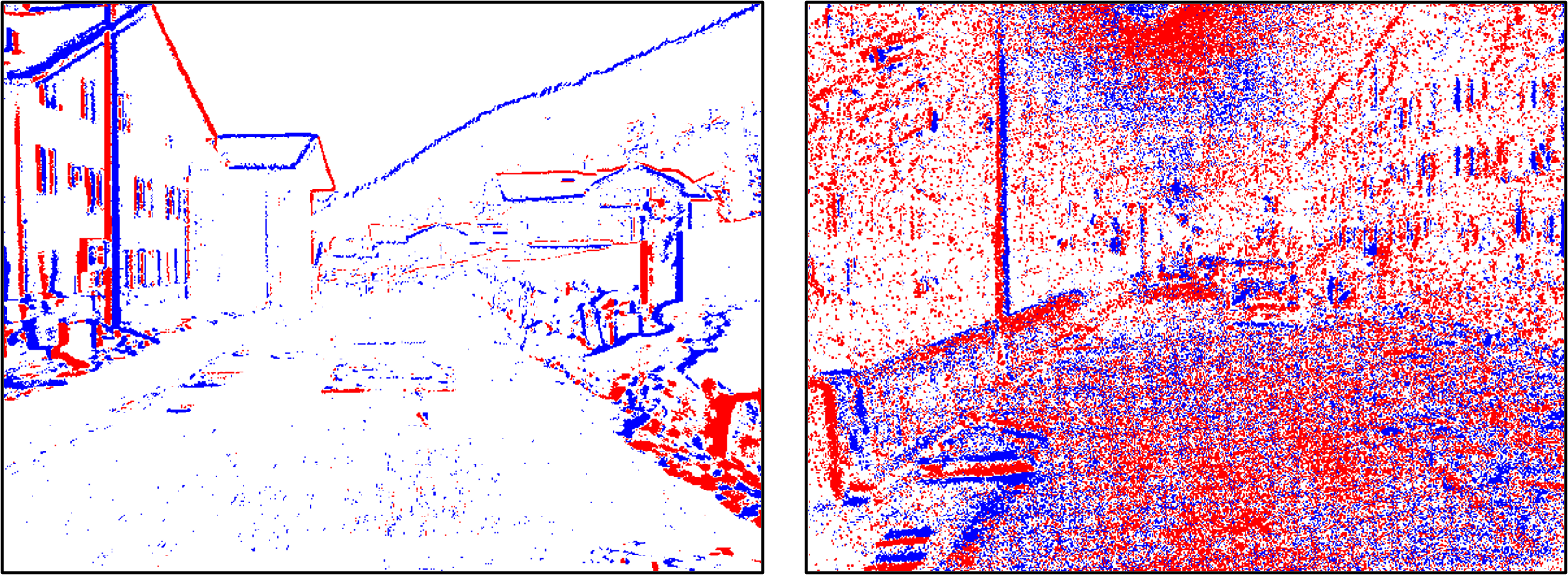}
  \caption{Event frames from \textit{interlaken\_01\_a} (daytime, left) and \textit{zurich\_city\_12\_a} (nighttime, right). Nighttime scenes exhibit significantly higher noise levels compared to well-lit daytime scenes.}
  \label{fig:eventnoise}
\end{figure}

However, the residual refiner offers limited improvement to the LTR model on the \textit{zurich\_city\_12\_a} sequence, likely due to its high noise levels. Fig.~\ref{fig:eventnoise} illustrates event frames from \textit{zurich\_city\_12\_a} (nighttime) and \textit{interlaken\_01\_a} (daytime), highlighting their respective noise levels. While event cameras effectively capture visual information in low-light environments, they also generate significant noise, which reduces IWE contrast and impairs evaluation metrics. This intense noise can adversely affect training, especially in self-supervised approaches. Paredes et al.~\cite{paredes2023taming} suggest excluding nighttime scenes from training to mitigate such effects.

Fig.~\ref{fig:mainresult} also presents examples of the initial LTR optical flow and the residual flow predicted by ResFlow. Analysis of the residual flows reveals that they are primarily distributed in two regions: the image margins and the boundaries between the foreground and background. Both areas are associated with occlusions. At the image margins, occlusions occur due to the appearance or disappearance of pixels. Similarly, dynamic occlusions at the foreground-background boundaries arise as the foreground moves over the background, creating motion patterns that linear models cannot effectively capture.

Finally, we emphasize the distinction between the sources of supervision and evaluation in our approach. Unlike other methods~\cite{paredes2023taming, shiba2022multicm} that rely on IWE contrast loss~\cite{gallego2018unifying} or average timestamp loss~\cite{zhu2018evflownet}, which inherently align with the FWL metric, our approach employs LTR-GT supervision, independent of FWL. This independence mitigates risks such as unfair evaluation and event collapse~\cite{shiba2022event}, commonly associated with contrast-maximum losses. By avoiding the incorporation of implicit priors during training, LTR-GT provides robust and fair supervision. Experimental results demonstrate that our method achieves superior FWL performance exclusively under LTR-GT supervision, highlighting its effectiveness and robustness.

\begin{figure}[t]
  \centering
  \includegraphics[width=0.99\linewidth]{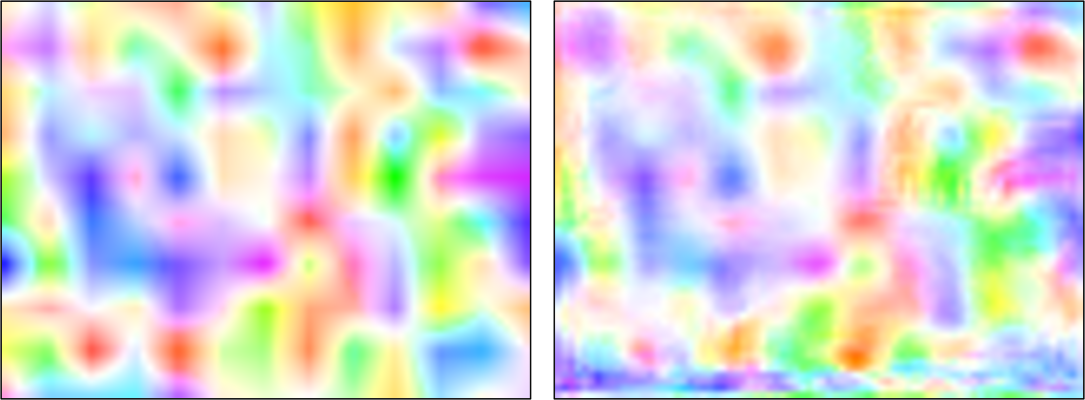}
  \caption{\textbf{Visualization of Local Perturbation Prediction.} The left figure illustrates the added regional noise $N_R$, while the right figure shows the corresponding prediction results $\bar{N}_R$. Both exhibit similar structural patterns, highlighting the effectiveness of ResFlow in correcting optical flow.}
  \label{fig:noisepred}
\end{figure}

\subsection{Ablation Study}\label{sec:ablation}

\begin{table}[t]
\centering
\caption{Ablation Study. $N_R$ and $N_W$ represent regional noise and white noise, respectively, and the numbers indicate the weight of noise.}
\begin{tabular}{@{}cccc@{}}
\toprule
\toprule
Method                    & Velocity Transformation & Noise Pattern       & FWL($\uparrow$)             \\ \midrule
LTR baseline~\cite{liu2023tma}                       & -     & -           & 2.074          \\ \midrule
\multirow{7}{*}{ResFlow}    & \XSolidBrush   & \XSolidBrush         & 2.080          \\
                          & \Checkmark     & \XSolidBrush         & 2.085          \\
                          & \Checkmark     & $N_R=0.1$       & 2.133          \\
                          & \Checkmark     & $N_R=0.3$       & \textbf{2.139} \\
                          & \Checkmark     & $N_R=0.5$       & 2.136          \\
                          & \XSolidBrush   & $N_R=0.3$       & 2.103          \\
                          & \Checkmark     & $N_W=0.3$ & 2.116          \\ \midrule
    Self-GT        & \Checkmark     & $N_R=0.3$ & 2.128          \\ \bottomrule\bottomrule
\end{tabular}
\label{tab:ablation}
\end{table}

\begin{figure}[t]
    \centering
        \includegraphics[width=1.\linewidth]{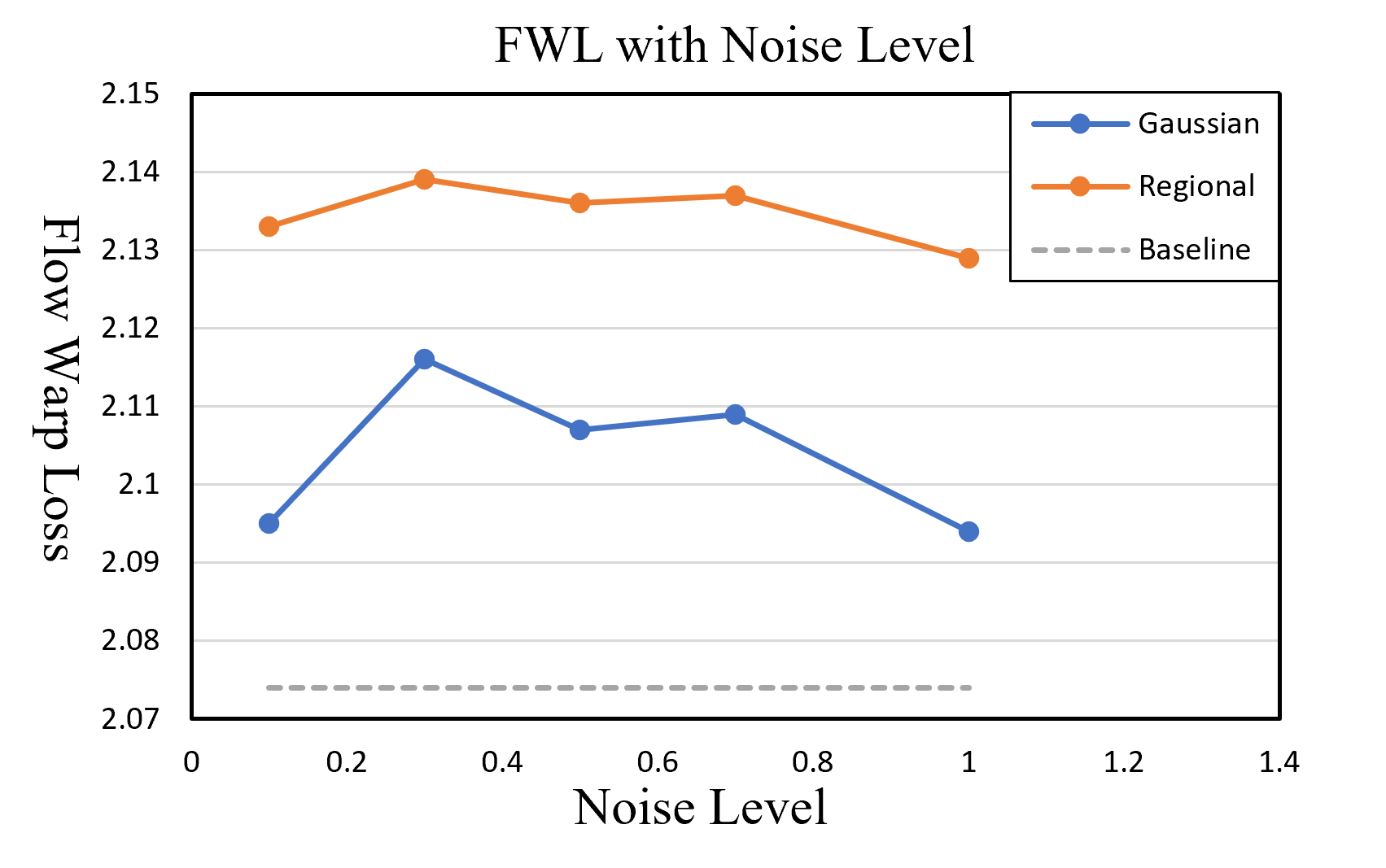}
        \vspace{-0.5cm}
        \caption{\textbf{Ablation study on noise pattern.} We evaluate the impact of injected noise on model performance (FWL metric) using two types of noise: Gaussian and region-based. Each data point represents a specific noise intensity. The baseline result without noise injection is also plotted for reference.}
    \label{fig:abl-noise}
\end{figure}

We conducted comprehensive ablation studies on the DSEC-Flow dataset, with results reported in Table~\ref{tab:ablation}. The ablation study primarily encompassed noise patterns, noise strength, velocity transformation, self-supervision, and inference frequency. In addition, we conducted a thorough ablation study and evaluation of the injected noise, with the results reported in Fig~\ref{fig:abl-noise}.

\vspace{0.5em}
\noindent\textbf{Noise Patterns.} As illustrated in Fig.~\ref{fig:resfconcept}, white noise encompasses both high- and low-frequency components, while regional noise contains more low-frequency components. We propose using regional noise as the perturbation due to its similarity to the residual flow pattern. To investigate the impact of different noise patterns, we performed separate training experiments using white noise and regional noise under consistent noise intensity. 
As shown in Table~\ref{tab:ablation} and Fig.~\ref{fig:abl-noise}, region-based noise consistently outperforms Gaussian noise in predicting HTR residual flow, supporting our hypothesis that its structured perturbation more closely aligns with the characteristics of residual flow patterns. In addition to improved accuracy, region-based noise also demonstrates greater robustness across a wide range of noise intensities.
This observation is further supported by Fig.~\ref{fig:mainresult}, which highlights the regional nature of residual flows. Additionally, Fig.~\ref{fig:noisepred} presents an example of noise correction by the residual refiner, demonstrating its ability to mitigate random perturbations and refine noisy optical flow.

\vspace{0.5em}
\noindent\textbf{Noise Strength.} 
The magnitude of the injected noise plays a critical role in balancing supervision quality and learning robustness. Excessive noise forces the residual refiner to rely on near-random perturbations, effectively shifting its role toward predicting the entire flow rather than refining it. On the other hand, overly weak noise fails to introduce sufficient perturbation, leading to limited supervisory signal and suboptimal learning. To identify an effective trade-off, we systematically varied the noise strength. As shown in Table~\ref{tab:ablation} and Fig.~\ref{fig:abl-noise}, a moderate noise weight of 0.3 achieves the best performance. Notably, compared to Gaussian noise, region-based noise demonstrates significantly greater robustness to changes in noise intensity. While the performance of Gaussian noise varies markedly with the noise level, region-based noise maintains stable and consistent improvements across a wide intensity range.

\begin{table}[t]
\centering
\caption{Performance at different inference frequencies. `Vox.', `Enc.', and `Corr.' indicate whether voxelization, feature encoding, and correlation are required, respectively. `Par.' and `T' represent model complexity and inference time, respectively.}
\begin{tabular}{@{}ccccccc@{}}
\toprule
\toprule
Frequency & Vox. & Enc. & Corr. & FWL    & Parameters & time(ms) \\ \midrule
10 Hz       &      --        &    --     &       --      & 2.074 & 6.9M   & 43       \\
50 Hz       &      \XSolidBrush        &     \XSolidBrush    &     \XSolidBrush        & 2.139 & 9.1M   & 49       \\
150 Hz      &      \XSolidBrush        &   \Checkmark      &   \Checkmark          & 2.143 & 9.1M   & 68       \\ \bottomrule \bottomrule
\end{tabular}
\label{tab:frequency}
\end{table}

\vspace{0.5em}
\noindent\textbf{Velocity Transformation.} To facilitate the adaptation of LTR supervision to HTR inference, we replaced HTR flow with scale-invariant velocity. With the unit time set to $T_{k+1}-T_k$, velocity transitions occur only during inference for intermediate flows. We conducted ablation studies on velocity transformation under both noisy and noise-free settings. As shown in Table~\ref{tab:ablation}, velocity transformations improve HTR residual estimation, particularly when combined with noise training.

\vspace{0.5em}
\noindent\textbf{Self-Supervision.} 
As detailed in Sec.~\ref{sec:method}, the training of the residual refiner can be extended to all in-domain data in a self-supervised manner if the endpoint error of the LTR estimation is low. In this setup, self-supervised training enables the network to refine residual predictions by learning to correct random disturbances. To validate this approach, we used LTR estimations as ground truth for training. Results in Table~\ref{tab:ablation} demonstrate the effectiveness of self-supervision. Notably, the self-supervised training shows significant potential, as it is conducted exclusively on in-domain data of equal size. These results highlight the critical role of the introduced noise, which significantly enhances performance even when the residual stream remains zero.

\vspace{0.5em}
\noindent\textbf{Inference Frequency.} To balance computational complexity, the constructed temporally dense cost volumes include five intermediate moments, enabling 50 Hz residual estimation with minimal overhead and eliminating the need to recompute cost volumes. The proposed residual refiner can estimate residual flows for any intermediate moment. To evaluate its performance, we tested residual flows up to 150 Hz, with results shown in Table~\ref{tab:frequency}. Notably, higher-frequency predictions do not require retraining, only the computation of cost volumes at intermediate moments. As shown in Table~\ref{tab:frequency}, 150 Hz ResFlow incurs approximately a 25 ms increase in computational cost, with a performance improvement of less than 0.01 compared to 50 Hz.

\revise{Experiments show that our residual-based framework effectively captures nonlinear motion and supports accurate HTR flow estimation. Key components such as velocity transformation, region-based noise, and self-supervision significantly enhance performance and robustness.}

\section{Conclusion}\label{sec:conclusion}
We have presented a novel residual paradigm for HTR optical flow estimation, enabling fast and accurate HTR motion estimation based on LTR linear trajectory. By decoupling LTR flow from HTR estimation, the residual paradigm mitigates the effects of event sparsity and seamlessly integrates with any LTR algorithms. Additionally, we present a regional noise strategy to facilitate the adaptation from LTR supervision to HTR inference, where regional noise effectively emulates the residual flow patterns. Residual prediction is enhanced by training the network to correct random disturbances. The noise-based strategy improves HTR performance and enables in-domain self-supervision. 
In the future, we aim to reduce reliance on LTR ground truth by exploring more general self-supervised learning strategies.
By predicting residual flow, our method addresses the trade-off between accuracy and frequency in HTR prediction and supports LTR supervision. 
Furthermore, our framework serves as a solid foundation for broader applications such as frame interpolation and high-precision motion estimation.
Comprehensive experiments on real-world datasets validate the effectiveness and superiority of the proposed approach.

\bibliographystyle{IEEEtran}
\bibliography{main}

\begin{thebibliography}{10}
\providecommand{\url}[1]{#1}
\csname url@samestyle\endcsname
\providecommand{\newblock}{\relax}
\providecommand{\bibinfo}[2]{#2}
\providecommand{\BIBentrySTDinterwordspacing}{\spaceskip=0pt\relax}
\providecommand{\BIBentryALTinterwordstretchfactor}{4}
\providecommand{\BIBentryALTinterwordspacing}{\spaceskip=\fontdimen2\font plus
\BIBentryALTinterwordstretchfactor\fontdimen3\font minus \fontdimen4\font\relax}
\providecommand{\BIBforeignlanguage}[2]{{%
\expandafter\ifx\csname l@#1\endcsname\relax
\typeout{** WARNING: IEEEtran.bst: No hyphenation pattern has been}%
\typeout{** loaded for the language `#1'. Using the pattern for}%
\typeout{** the default language instead.}%
\else
\language=\csname l@#1\endcsname
\fi
#2}}
\providecommand{\BIBdecl}{\relax}
\BIBdecl

\bibitem{dosovitskiy2015flownet}
A.~Dosovitskiy, P.~Fischer, E.~Ilg, P.~Hausser, C.~Hazirbas, V.~Golkov, P.~Van Der~Smagt, D.~Cremers, and T.~Brox, ``Flownet: Learning optical flow with convolutional networks,'' in \emph{Proceedings of the IEEE International Conference on Computer Vision}, 2015, pp. 2758--2766.

\bibitem{sun2018pwcnet}
D.~Sun, X.~Yang, M.-Y. Liu, and J.~Kautz, ``Pwc-net: Cnns for optical flow using pyramid, warping, and cost volume,'' in \emph{Proceedings of the IEEE Conference on Computer Vision and Pattern Recognition}, 2018, pp. 8934--8943.

\bibitem{teed2020raft}
Z.~Teed and J.~Deng, ``Raft: Recurrent all-pairs field transforms for optical flow,'' in \emph{European Conference on Computer Vision}.\hskip 1em plus 0.5em minus 0.4em\relax Springer, 2020, pp. 402--419.

\bibitem{jiang2021gma}
S.~Jiang, D.~Campbell, Y.~Lu, H.~Li, and R.~Hartley, ``Learning to estimate hidden motions with global motion aggregation,'' in \emph{Proceedings of the IEEE/CVF International Conference on Computer Vision}, 2021, pp. 9772--9781.

\bibitem{gallego2020event}
G.~Gallego, T.~Delbr{\"u}ck, G.~Orchard, C.~Bartolozzi, B.~Taba, A.~Censi, S.~Leutenegger, A.~J. Davison, J.~Conradt, K.~Daniilidis \emph{et~al.}, ``Event-based vision: A survey,'' \emph{IEEE Transactions on Pattern Analysis and Machine Intelligence}, vol.~44, no.~1, pp. 154--180, 2020.

\bibitem{liu2023motiondeblur}
Z.~Liu, J.~Wu, G.~Shi, W.~Yang, W.~Dong, and Q.~Zhao, ``Motion-oriented hybrid spiking neural networks for event-based motion deblurring,'' \emph{IEEE Transactions on Circuits and Systems for Video Technology}, 2023.

\bibitem{zhu2023learningdeblur}
Q.~Zhu, N.~Zheng, J.~Huang, M.~Zhou, J.~Zhang, and F.~Zhao, ``Learning spatio-temporal sharpness map for video deblurring,'' \emph{IEEE Transactions on Circuits and Systems for Video Technology}, 2023.

\bibitem{zhu2022learning}
Z.~Zhu, J.~Hou, and X.~Lyu, ``Learning graph-embedded key-event back-tracing for object tracking in event clouds,'' \emph{Advances in Neural Information Processing Systems}, vol.~35, pp. 7462--7476, 2022.

\bibitem{chen2025event}
N.~Chen, C.~Zhang, W.~An, L.~Wang, M.~Li, and Q.~Ling, ``Event-based motion deblurring with blur-aware reconstruction filter,'' \emph{IEEE Transactions on Circuits and Systems for Video Technology}, 2025.

\bibitem{nie2020highhtr}
K.~Nie, X.~Shi, S.~Cheng, Z.~Gao, and J.~Xu, ``High frame rate video reconstruction and deblurring based on dynamic and active pixel vision image sensor,'' \emph{IEEE Transactions on Circuits and Systems for Video Technology}, vol.~31, no.~8, pp. 2938--2952, 2020.

\bibitem{yu2025event}
J.~Yu, X.~Lu, L.~Guo, C.~Wang, G.~Li, and J.~Qian, ``Event-based video reconstruction via spatial-temporal heterogeneous spiking neural network,'' \emph{IEEE Transactions on Circuits and Systems for Video Technology}, 2025.

\bibitem{wang2023visevent}
X.~Wang, J.~Li, L.~Zhu, Z.~Zhang, Z.~Chen, X.~Li, Y.~Wang, Y.~Tian, and F.~Wu, ``Visevent: Reliable object tracking via collaboration of frame and event flows,'' \emph{IEEE Transactions on Cybernetics}, 2023.

\bibitem{zhu2024crsot}
Y.~Zhu, X.~Wang, C.~Li, B.~Jiang, L.~Zhu, Z.~Huang, Y.~Tian, and J.~Tang, ``Crsot: Cross-resolution object tracking using unaligned frame and event cameras,'' \emph{arXiv preprint arXiv:2401.02826}, 2024.

\bibitem{chen2022ecsnetstf}
Z.~Chen, J.~Wu, J.~Hou, L.~Li, W.~Dong, and G.~Shi, ``Ecsnet: Spatio-temporal feature learning for event camera,'' \emph{IEEE Transactions on Circuits and Systems for Video Technology}, vol.~33, no.~2, pp. 701--712, 2022.

\bibitem{chen2024segment}
Z.~Chen, Z.~Zhu, Y.~Zhang, J.~Hou, G.~Shi, and J.~Wu, ``Segment any event streams via weighted adaptation of pivotal tokens,'' in \emph{Proceedings of the IEEE/CVF Conference on Computer Vision and Pattern Recognition}, 2024, pp. 3890--3900.

\bibitem{xie2024eventstf}
B.~Xie, Y.~Deng, Z.~Shao, Q.~Xu, and Y.~Li, ``Event voxel set transformer for spatiotemporal representation learning on event streams,'' \emph{IEEE Transactions on Circuits and Systems for Video Technology}, 2024.

\bibitem{zhu2024continuous}
L.~Zhu, X.~Chen, L.~Wang, X.~Wang, Y.~Tian, and H.~Huang, ``Continuous-time object segmentation using high temporal resolution event camera,'' \emph{IEEE Transactions on Pattern Analysis and Machine Intelligence}, 2024.

\bibitem{deng2021mvfll}
Y.~Deng, H.~Chen, and Y.~Li, ``Mvf-net: A multi-view fusion network for event-based object classification,'' \emph{IEEE Transactions on Circuits and Systems for Video Technology}, vol.~32, no.~12, pp. 8275--8284, 2021.

\bibitem{liu2024video}
Y.~Liu, Y.~Deng, H.~Chen, and Z.~Yang, ``Video frame interpolation via direct synthesis with the event-based reference,'' in \emph{Proceedings of the IEEE/CVF Conference on Computer Vision and Pattern Recognition}, 2024, pp. 8477--8487.

\bibitem{liu2022edflow}
M.~Liu and T.~Delbruck, ``Edflow: Event driven optical flow camera with keypoint detection and adaptive block matching,'' \emph{IEEE Transactions on Circuits and Systems for Video Technology}, vol.~32, no.~9, pp. 5776--5789, 2022.

\bibitem{zhu2023cross}
Z.~Zhu, J.~Hou, and D.~O. Wu, ``Cross-modal orthogonal high-rank augmentation for rgb-event transformer-trackers,'' in \emph{Proceedings of the IEEE/CVF International Conference on Computer Vision}, 2023, pp. 22\,045--22\,055.

\bibitem{wu2024motion}
S.~Wu, Z.~Zhu, J.~Hou, G.~Shi, and J.~Wu, ``E-motion: Future motion simulation via event sequence diffusion,'' in \emph{Advances in Neural Information Processing Systems}, 2024.

\bibitem{chen2024crossei}
Z.~Chen, J.~Wu, W.~Dong, L.~Li, and G.~Shi, ``Crossei: Boosting motion-oriented object tracking with an event camera,'' \emph{IEEE Transactions on Image Processing}, 2024.

\bibitem{zhu2025spatio}
Y.~Zhu, Y.~Gao, T.~Ding, X.~Liu, W.~Yang, and T.~Zhang, ``Spatio-temporal pyramid keypoint detection with event cameras,'' \emph{IEEE Transactions on Circuits and Systems for Video Technology}, 2025.

\bibitem{shi2024polarity}
C.~Shi, B.~Wei, X.~Wang, H.~Liu, Y.~Zhang, W.~Li, N.~Song, and J.~Jin, ``Polarity-focused denoising for event cameras,'' \emph{IEEE Transactions on Circuits and Systems for Video Technology}, 2024.

\bibitem{liu2024event}
X.~Liu, J.~Li, J.~Shi, X.~Fan, Y.~Tian, and D.~Zhao, ``Event-based monocular depth estimation with recurrent transformers,'' \emph{IEEE Transactions on Circuits and Systems for Video Technology}, 2024.

\bibitem{khan2022non}
A.~T. Khan, X.~Cao, I.~Brajevic, P.~S. Stanimirovic, V.~N. Katsikis, and S.~Li, ``Non-linear activated beetle antennae search: A novel technique for non-convex tax-aware portfolio optimization problem,'' \emph{Expert Systems with Applications}, vol. 197, p. 116631, 2022.

\bibitem{cao2025decomposition}
X.~Cao, J.~Lou, B.~Liao, C.~Peng, X.~Pu, A.~T. Khan, D.~T. Pham, and S.~Li, ``Decomposition based neural dynamics for portfolio management with tradeoffs of risks and profits under transaction costs,'' \emph{Neural Networks}, vol. 184, p. 107090, 2025.

\bibitem{khan2021quantum}
A.~T. Khan, X.~Cao, S.~Li, B.~Hu, and V.~N. Katsikis, ``Quantum beetle antennae search: a novel technique for the constrained portfolio optimization problem,'' \emph{Science China Information Sciences}, vol.~64, pp. 1--14, 2021.

\bibitem{moshayedi2023designing}
A.~J. Moshayedi, N.~M.~I. Uddin, A.~S. Khan, J.~Zhu, and M.~Emadi~Andani, ``Designing and developing a vision-based system to investigate the emotional effects of news on short sleep at noon: an experimental case study,'' \emph{Sensors}, vol.~23, no.~20, p. 8422, 2023.

\bibitem{moshayedi2024design}
A.~J. Moshayedi, A.~S. Roy, L.~Liao, A.~S. Khan, A.~Kolahdooz, and A.~Eftekhari, ``Design and development of foodiebot robot: From simulation to design,'' \emph{IEEE access}, vol.~12, pp. 36\,148--36\,172, 2024.

\bibitem{moshayedi2024evaluating}
A.~J. Moshayedi, Y.~Xie, M.~Sharifdoust, and A.~S. Khan, ``Evaluating omni robot navigation with slam in coppeliasim: Hemangiomas and nonhomogeneous paths,'' \emph{transformation}, vol.~1, p.~v2, 2024.

\bibitem{r2r1}
E.~Aghajari and A.~A. AbdulRahim, ``Prediction of short circuit current of wind turbines based on artificial neural network model,'' \emph{EAI Endorsed Trans. AI Robot}, vol.~3, 2024.

\bibitem{r2r2}
Z.~Li, S.~Li, O.~O. Bamasag, A.~Alhothali, and X.~Luo, ``Diversified regularization enhanced training for effective manipulator calibration,'' \emph{IEEE Transactions on Neural Networks and Learning Systems}, vol.~34, no.~11, pp. 8778--8790, 2023.

\bibitem{r2r3}
X.~Cao, Y.~Yang, S.~Li, P.~S. Stanimirovi{\'c}, and V.~N. Katsikis, ``Artificial neural dynamics for portfolio allocation: An optimization perspective,'' \emph{IEEE Transactions on Systems, Man, and Cybernetics: Systems}, 2024.

\bibitem{r2r4}
Z.~Li, S.~Li, A.~Francis, and X.~Luo, ``A novel calibration system for robot arm via an open dataset and a learning perspective,'' \emph{IEEE Transactions on Circuits and Systems II: Express Briefs}, vol.~69, no.~12, pp. 5169--5173, 2022.

\bibitem{paredes2023taming}
F.~Paredes-Vall{\'e}s, K.~Y. Scheper, C.~De~Wagter, and G.~C. De~Croon, ``Taming contrast maximization for learning sequential, low-latency, event-based optical flow,'' in \emph{Proceedings of the IEEE/CVF International Conference on Computer Vision}, 2023, pp. 9695--9705.

\bibitem{gallego2018unifying}
G.~Gallego, H.~Rebecq, and D.~Scaramuzza, ``A unifying contrast maximization framework for event cameras, with applications to motion, depth, and optical flow estimation,'' in \emph{Proceedings of the IEEE Conference on Computer Vision and Pattern Recognition}, 2018, pp. 3867--3876.

\bibitem{zhu2018evflownet}
A.~Z. Zhu, L.~Yuan, K.~Chaney, and K.~Daniilidis, ``Ev-flownet: Self-supervised optical flow estimation for event-based cameras,'' \emph{arXiv preprint arXiv:1802.06898}, 2018.

\bibitem{ye2023towards}
Y.~Ye, H.~Shi, K.~Yang, Z.~Wang, X.~Yin, Y.~Lin, M.~Liu, Y.~Wang, and K.~Wang, ``Towards anytime optical flow estimation with event cameras,'' \emph{arXiv preprint arXiv:2307.05033}, 2023.

\bibitem{ponghiran2023sequential}
W.~Ponghiran, C.~M. Liyanagedera, and K.~Roy, ``Event-based temporally dense optical flow estimation with sequential learning,'' in \emph{Proceedings of the IEEE/CVF International Conference on Computer Vision}, 2023, pp. 9827--9836.

\bibitem{hui2018liteflownet}
T.-W. Hui, X.~Tang, and C.~C. Loy, ``Liteflownet: A lightweight convolutional neural network for optical flow estimation,'' in \emph{Proceedings of the IEEE Conference on Computer Vision and Pattern Recognition}, 2018, pp. 8981--8989.

\bibitem{xu2022gmflow}
H.~Xu, J.~Zhang, J.~Cai, H.~Rezatofighi, and D.~Tao, ``Gmflow: Learning optical flow via global matching,'' in \emph{Proceedings of the IEEE/CVF Conference on Computer Vision and Pattern Recognition}, 2022, pp. 8121--8130.

\bibitem{huang2022flowformer}
Z.~Huang, X.~Shi, C.~Zhang, Q.~Wang, K.~C. Cheung, H.~Qin, J.~Dai, and H.~Li, ``Flowformer: A transformer architecture for optical flow,'' in \emph{European Conference on Computer Vision}.\hskip 1em plus 0.5em minus 0.4em\relax Springer, 2022, pp. 668--685.

\bibitem{saxena2024surprising}
S.~Saxena, C.~Herrmann, J.~Hur, A.~Kar, M.~Norouzi, D.~Sun, and D.~J. Fleet, ``The surprising effectiveness of diffusion models for optical flow and monocular depth estimation,'' \emph{Advances in Neural Information Processing Systems}, vol.~36, 2024.

\bibitem{gehrig2021eraft}
M.~Gehrig, M.~Millh{\"a}usler, D.~Gehrig, and D.~Scaramuzza, ``E-raft: Dense optical flow from event cameras,'' in \emph{2021 International Conference on 3D Vision (3DV)}.\hskip 1em plus 0.5em minus 0.4em\relax IEEE, 2021, pp. 197--206.

\bibitem{wu2024idnet}
Y.~Wu, F.~Paredes-Vall{\'e}s, and G.~C. De~Croon, ``Lightweight event-based optical flow estimation via iterative deblurring,'' in \emph{2024 IEEE International Conference on Robotics and Automation (ICRA)}.\hskip 1em plus 0.5em minus 0.4em\relax IEEE, 2024, pp. 14\,708--14\,715.

\bibitem{yang2025ecddp}
Y.~Yang, L.~Pan, and L.~Liu, ``Event camera data dense pre-training,'' in \emph{European Conference on Computer Vision}.\hskip 1em plus 0.5em minus 0.4em\relax Springer, 2025, pp. 292--310.

\bibitem{liu2023tma}
H.~Liu, G.~Chen, S.~Qu, Y.~Zhang, Z.~Li, A.~Knoll, and C.~Jiang, ``Tma: Temporal motion aggregation for event-based optical flow,'' in \emph{Proceedings of the IEEE/CVF International Conference on Computer Vision}, 2023, pp. 9685--9694.

\bibitem{shiba2022multicm}
S.~Shiba, Y.~Aoki, and G.~Gallego, ``Secrets of event-based optical flow,'' in \emph{European Conference on Computer Vision}.\hskip 1em plus 0.5em minus 0.4em\relax Springer, 2022, pp. 628--645.

\bibitem{shiba2024secrets}
S.~Shiba, Y.~Klose, Y.~Aoki, and G.~Gallego, ``Secrets of event-based optical flow, depth and ego-motion estimation by contrast maximization,'' \emph{IEEE Transactions on Pattern Analysis and Machine Intelligence}, 2024.

\bibitem{hagenaars2021nips}
J.~Hagenaars, F.~Paredes-Vall{\'e}s, and G.~De~Croon, ``Self-supervised learning of event-based optical flow with spiking neural networks,'' \emph{Advances in Neural Information Processing Systems}, vol.~34, pp. 7167--7179, 2021.

\bibitem{hamann2024motion}
F.~Hamann, Z.~Wang, I.~Asmanis, K.~Chaney, G.~Gallego, and K.~Daniilidis, ``Motion-prior contrast maximization for dense continuous-time motion estimation,'' in \emph{European Conference on Computer Vision}.\hskip 1em plus 0.5em minus 0.4em\relax Springer, 2025, pp. 18--37.

\bibitem{wan2022dceiflow}
Z.~Wan, Y.~Dai, and Y.~Mao, ``Learning dense and continuous optical flow from an event camera,'' \emph{IEEE Transactions on Image Processing}, vol.~31, pp. 7237--7251, 2022.

\bibitem{gehrig2024bflow}
M.~Gehrig, M.~Muglikar, and D.~Scaramuzza, ``Dense continuous-time optical flow from event cameras,'' \emph{IEEE Transactions on Pattern Analysis and Machine Intelligence}, 2024.

\bibitem{gehrig2021dsec}
M.~Gehrig, W.~Aarents, D.~Gehrig, and D.~Scaramuzza, ``Dsec: A stereo event camera dataset for driving scenarios,'' \emph{IEEE Robotics and Automation Letters}, vol.~6, no.~3, pp. 4947--4954, 2021.

\bibitem{shiba2022event}
S.~Shiba, Y.~Aoki, and G.~Gallego, ``Event collapse in contrast maximization frameworks,'' \emph{Sensors}, vol.~22, no.~14, p. 5190, 2022.

\end{thebibliography}

\end{document}